\documentclass{article}

\usepackage[preprint]{neurips_2026}
\usepackage[utf8]{inputenc}
\usepackage[T1]{fontenc}
\usepackage{microtype}
\usepackage{graphicx}
\usepackage{url}
\usepackage{booktabs}
\usepackage{amsmath}
\usepackage{amssymb}
\usepackage{amsfonts}
\usepackage{nicefrac}
\usepackage{algorithm}
\usepackage{algpseudocode}
\usepackage{enumitem}
\usepackage[most]{tcolorbox}
\usepackage{tikz}
\usepackage{xspace}
\usepackage{hyperref}
\hypersetup{colorlinks=true,citecolor=blue,linkcolor=blue,urlcolor=blue}

\newcommand{\name}{EMBER\xspace}
\definecolor{promptboxbg}{RGB}{240,245,253}
\definecolor{promptboxframe}{RGB}{47,88,166}
\newtcblisting{PromptBlock}{
  enhanced,
  breakable,
  listing only,
  listing engine=listings,
  colback=promptboxbg,
  colframe=promptboxframe,
  boxrule=1.1pt,
  arc=9pt,
  outer arc=9pt,
  boxsep=5pt,
  left=6pt,
  right=6pt,
  top=5pt,
  bottom=5pt,
  before skip=0.8em,
  after skip=1.0em,
  listing options={
    basicstyle=\ttfamily\small,
    breaklines=true,
    breakatwhitespace=false,
    columns=fullflexible,
    keepspaces=true
  }
}
\newtcblisting{CodeBlock}{
  enhanced,
  breakable,
  listing only,
  listing engine=listings,
  colback=promptboxbg,
  colframe=promptboxframe,
  boxrule=1.1pt,
  arc=9pt,
  outer arc=9pt,
  boxsep=5pt,
  left=6pt,
  right=6pt,
  top=5pt,
  bottom=5pt,
  before skip=0.8em,
  after skip=1.0em,
  listing options={
    basicstyle=\ttfamily\small,
    breaklines=true,
    breakatwhitespace=false,
    columns=fullflexible,
    keepspaces=true
  }
}

\title{EMBER: Efficient Memory via Budgeted Evidence \\Retention for Long-Horizon Agents}

\author{
  Yilong Li and Suman Banerjee \\
  University of Wisconsin--Madison \\
  \texttt{\{yilong,suman\}@cs.wisc.edu}
  \And
  Tong Che \\
  NVIDIA Research \\
  \texttt{tongc@nvidia.com}
}
\vspace{-0.7ex}

\begin{document}
\raggedbottom

\maketitle

\begin{abstract}
Long-horizon agents can archive large histories, yet answering future requests
still consumes retrieval and context tokens. If retained memory omits
answer-relevant evidence, the system must revisit more of the raw history. We
study \emph{budgeted evidence survival}: selecting, before the query is known,
the source evidence that should remain recoverable under a fixed retained
source-evidence budget. We instantiate this problem as \emph{Budgeted Pre-Query
Retention}, where memory is written during ingestion and later read without
access to the full raw stream. We introduce \name, a learned retention policy
that constructs a compact, source-backed evidence state. \name stores evidence
capsules: verbatim source excerpts paired with retrieval keys and update
metadata, preserving both grounding and read-time access. Post-query outcome
feedback trains the writer across the ingestion--retrieval--answer chain. On
LongMemEval-RR, our LongMemEval-derived retained-evidence protocol, \name-14B
reaches 0.3017 F1 at the 8192-token retained-evidence comparison point, compared
with 0.1765 for the strongest non-\name budgeted baseline. Across retained
source-evidence budgets, \name improves F1, Retain-Recall, and Read-Recall,
indicating that long-horizon memory depends on evidence retained within the
budget, not simply on rereading larger histories.

\end{abstract}

\vspace{-0.65ex}
\section{Introduction}
\label{sec:introduction}
\vspace{-0.65ex}
Long-horizon agents cannot treat memory as a free raw-log archive. Bytes can be
stored; usable context is consumed at read time. Documents, conversations,
files, and tool traces must still be indexed, retrieved, and read through model
context. When a memory write misses useful evidence, the system must broaden
search or send more history back through the model.

Full-log RAG postpones the choice of what matters because the corpus remains
available at read time. Query-visible memory agents see the target task before
deciding what to rewrite or route, as in MemAgent~\citep{yu2026memagent} and
learned memory-operation systems optimized for downstream
QA~\citep{yan2025memoryr1,wang2025memalpha}. Persistent agents often have neither
luxury: they write memory while the future request is unknown, and the full raw
stream cannot remain query-time usable.
We study this as \emph{budgeted evidence survival}. Budgeted Pre-Query Retention
is the protocol: the agent builds memory during ingestion, keeps only a fixed
amount of source evidence, denoted $B_{\mathrm{ret}}$ below, and later answers
without returning to the full stream. Pre-query writing fixes when the decision
is made; the objective is to allocate the budget to evidence the reader can
recover and use. Efficiency here means token efficiency: the same
$B_{\mathrm{ret}}$ should preserve more evidence useful for future answers.

\name learns this choice directly. It stores compact evidence capsules: source
excerpts with retrieval keys that make them findable after the query arrives.
On LongMemEval-RR, which keeps the LongMemEval histories, questions, and answers
fixed while changing the resource regime~\begin{NoHyper}\citep{wu2025longmemeval}\end{NoHyper}, \name-14B
reaches 0.3017 F1 at the shared 8192-token budget, compared with 0.1765 for the
strongest budgeted non-\name baseline.

\vspace{-0.45ex}
\begin{figure}[H]
  \centering
  \includegraphics[width=0.90\textwidth]{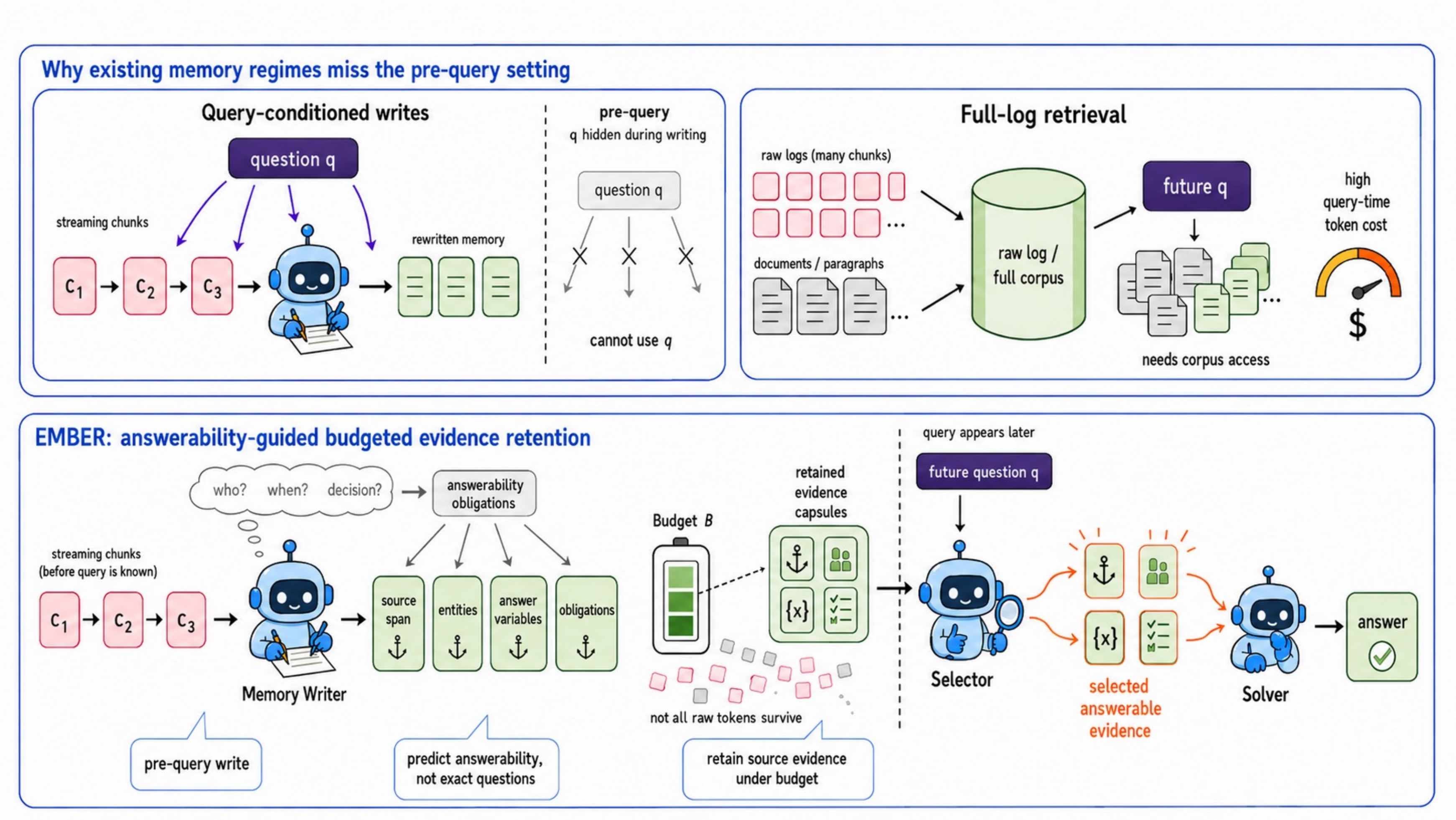}
  \caption{\name improves evidence survival under $B_{\mathrm{ret}}$:
  answerability probes select source spans, and retrieval keys keep them
  searchable at read time.}
  \label{fig:online-memory-management}
  \vspace{-0.35ex}
\end{figure}
\vspace{-0.45ex}

We report Retain-Recall, Read-Recall, and F1 so the main comparison separates
write failures, read failures, and final answer quality.

\vspace{-0.45ex}
\begin{figure}[htb]
  \centering
  \includegraphics[width=0.98\textwidth]{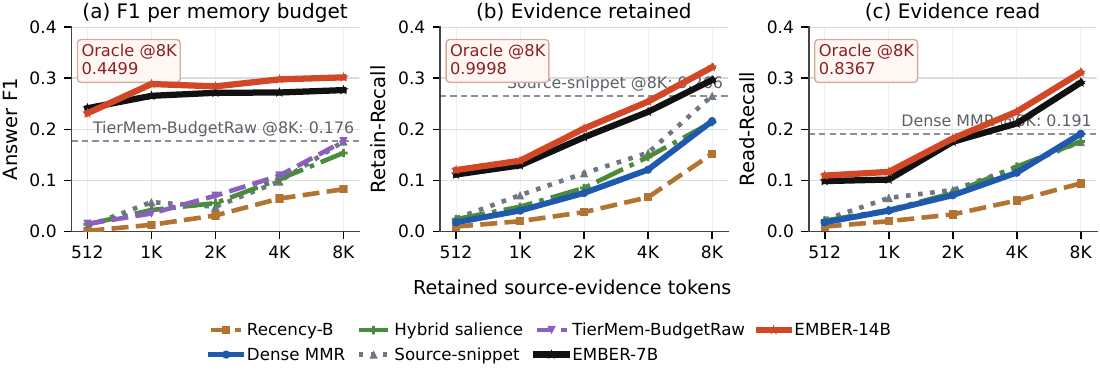}
\caption{\textbf{Memory-token efficiency frontier.} We vary $B_{\mathrm{ret}}$
  and report F1, Retain-Recall, and Read-Recall. The main table fixes
  $B=8192$ for the cross-method comparison; this figure shows the full budget
  frontier. Dashed references mark the best non-\name budgeted value at
  $B=8192$ in each panel; Oracle annotations show the matched $B=8192$ reference.}
  \label{fig:budget-frontier}
  \vspace{-0.35ex}
\end{figure}
\vspace{-0.45ex}

\begin{table}[htb]
  \centering
  \caption{Main results on LongMemEval-RR at $B_{\mathrm{ret}}=8192$ and
  top-$k=10$. F1 is reported as mean $\pm$ half-width of the 95\%
  nonparametric bootstrap confidence interval over 500 examples. Budgeted rows
  write memory before the query is known and have no
  raw-log access at read time; full-log and query-visible rows are reference
  categories. Retained-memory and full-log rows use a GPT-4o reader after
  retrieval; for \name, 7B/14B denotes the memory-policy writer backbone.
  Oracle retention packs gold-annotated evidence into the same budget.}
  \label{tab:longmemeval-rr-main}
  
  \small
  \setlength{\tabcolsep}{2.6pt}
  \begin{NoHyper}
  \resizebox{\linewidth}{!}{%
  \begin{tabular}{@{}llllccc@{}}
    \toprule
    \textbf{Method} &
    \textbf{Query visible at write time?} & \textbf{Full raw-log access?} & \textbf{Source-evidence budget} &
    \textbf{Retain-Recall $\uparrow$} & \textbf{Read-Recall $\uparrow$} & \textbf{F1 $\uparrow$} \\
    \midrule
    \multicolumn{7}{@{}l}{\textit{\textbf{Full-log references}}} \\
    Full raw-log RAG & No & Yes & full log & 1.0000 & 0.6082 & $0.3645{\pm}0.042$ \\
    TierMem with full raw logs & No & Yes & full log & 1.0000 & 0.6152 & $0.4541{\pm}0.044$ \\
    \midrule
    \multicolumn{7}{@{}l}{\textit{\textbf{Query-visible reference}}} \\
    MemAgent-7B & Yes & No & native & N/A & N/A & $0.4633{\pm}0.044$ \\
    \midrule
    \multicolumn{7}{@{}l}{\textit{\textbf{Upper Bound}}} \\
    Oracle retention & No & No & 8192 & 0.9998 & 0.8367 & $0.4499{\pm}0.044$ \\
    \midrule
    \multicolumn{7}{@{}l}{\textit{\textbf{Budgeted evidence-survival baselines}}} \\
    Random-$B$ & No & No & 8192 & 0.0612 & 0.0425 & $0.0350{\pm}0.016$ \\
    Recency-$B$ & No & No & 8192 & 0.1521 & 0.0943 & $0.0827{\pm}0.024$ \\
    Reservoir-$B$ & No & No & 8192 & 0.0753 & 0.0615 & $0.0498{\pm}0.019$ \\
    Hybrid salience & No & No & 8192 & 0.2140 & 0.1763 & $0.1538{\pm}0.032$ \\
    Summary-only & No & No & $B$-equiv. & N/A & N/A & $0.1156{\pm}0.028$ \\
    MemAgent-7B (adapted) & No & No & 8192 & N/A & N/A & $0.1311{\pm}0.030$\\
    Source-snippet heuristic & No & No & 8192 & 0.2657 & 0.1754 & $0.1763{\pm}0.033$ \\
    Generated-query indexing & No & No & 8192 & 0.2021 & 0.1423 & $0.1205{\pm}0.029$ \\
    TierMem with budgeted raw fallback & No & No & 8192 & 0.1246 & 0.0997 & $0.1765{\pm}0.033$ \\
    Memory-R1 (adapted) & No & No & 8192 & 0.1714 & 0.1464 & $0.1565{\pm}0.032$ \\
    \midrule
    \multicolumn{7}{@{}l}{\textit{\textbf{\name}}} \\
    \textbf{\name-7B writer} (Qwen2.5-7B) & No & No & 8192 & \textbf{0.2966} & \textbf{0.2915} & $\mathbf{0.2768}{\pm}0.039$ \\
    \textbf{\name-14B writer} (Qwen2.5-14B) & No & No & 8192 & \textbf{0.3215} & \textbf{0.3112} & $\mathbf{0.3017}{\pm}0.040$ \\
    \bottomrule
  \end{tabular}
  }
  \end{NoHyper}
  \vspace{-0.35ex}
  \footnotesize
  \begin{NoHyper}
  \parbox{0.98\linewidth}{\textit{Notes.} Oracle retention packs annotated
  gold-evidence turns into the same $B_{\mathrm{ret}}$; it uses gold evidence
  labels during retention but no answer text at read time. Appendix
  Table~\ref{tab:longmemeval-rr-reader-check} gives extended reader diagnostics
  where logged; \name rows report only the synchronized F1/Retain/Read run used
  in the main comparison. Appendix~\ref{app:longmemeval-rr}
  defines the metrics formally.}
  \end{NoHyper}
  
\end{table}
\vspace{-0.65ex}

\begin{NoHyper}
Among budgeted evidence-survival baselines, TierMem with
budgeted raw fallback gives the top F1, essentially tied with the
source-snippet heuristic, while source snippets give the strongest
Retain-Recall. Summaries can help answer some questions even when they leave no
source-traceable evidence behind. \name-14B improves over both routes at the
common max-budget point: it keeps more gold evidence, brings more of it into the
reader, and reaches 0.3017 F1. Figure~\ref{fig:budget-frontier} extends the
comparison across budgets; at the same $B_{\mathrm{ret}}$, \name gets more gold
evidence through writing and reading. At the shared 8192-token point, oracle
retention reaches 0.4499 F1, leaving a measurable gap for better source
selection and read-time access.
\end{NoHyper}

\vspace{-0.65ex}
\paragraph{Contributions.}
Our contributions can be summarized as follows:
\begin{itemize}[leftmargin=1.5em]
  \item We formalize budgeted evidence survival as Budgeted Pre-Query Retention
  and instantiate it as LongMemEval-RR, with metrics that separate retention
  from read-time access.
  \vspace{-0.25ex}
  \item We introduce an answerability-guided retention method that stores source
  excerpts with retrieval keys, keeping generated summaries and future-question
  guesses out of the retained object.
  \vspace{-0.25ex}
  \item We train the retention policy with answer-time outcome feedback,
  enabling the agent to preserve evidence that remains useful under
  $B_{\mathrm{ret}}$.
  \vspace{-0.35ex}
\end{itemize}

Together, these pieces make the write step accountable: before any query is
observed, \name must choose the source-backed capsules that will carry evidence
into the read-time budget.

\vspace{-0.75ex}
\section{Problem Formulation}
\label{sec:problem}
\vspace{-0.75ex}

An episode consists of a stream, a future query hidden during ingestion, a
retained memory state, and an answer produced from that memory under budget.

\vspace{-0.75ex}
\subsection{Streaming Episodes}
\label{sec:streaming-episodes}
\vspace{-0.65ex}
Each episode is drawn from a distribution $D$ and provides a stream
$c_{1:K}=(c_1,\ldots,c_K)$, a query $q$, and a target answer $y$. During
ingestion, the agent observes the stream in order and must retain memory without
access to $q$. After ingestion, $q$ becomes available, and the agent produces an
answer $\hat{y}$ from its active context and retained memory.

\vspace{-0.85ex}
\subsection{Pre-Query Online Memory}
\label{sec:prequery-online-memory}
\vspace{-0.75ex}
What makes the setting hard is the timing of query access. During ingestion, the
agent sees only the stream so far and its current memory state; the query $q$ is
still unknown, so a write cannot be tailored to it. Only at read time does $q$
arrive, and the agent may then retrieve or select from the retained state. Thus,
an online agent must commit memory before learning what will be asked.

\vspace{-0.85ex}
\subsection{Memory State}
\label{sec:memory-state}
\vspace{-0.75ex}
At step $t$, the agent maintains an active context $w_t$ and an external memory
state $S_t$. The active context is a fixed-size working buffer used for streaming
computation and write triggering. It does not grow with the stream or with the
budget, and it is not a second evidence store. The external memory $S_t$ is the
persistent state: a set of retained capsules that outlive the active context
across the episode. The experimental budget $B_{\mathrm{ret}}$ limits retained
source-excerpt tokens available after ingestion:
\[
  |S_K|_{\mathrm{tok}} \le B_{\mathrm{ret}} .
\]
Throughout, ``budget'' and the symbol $B_{\mathrm{ret}}$ both refer to this
retained source-evidence token cap unless noted. Retrieval metadata is serialized
and reported separately, not counted as answer-bearing source evidence.
The desired memory state is evidence-grounded: it should expose retrievable
handles for future queries while preserving source evidence that can be returned
to the reader.

\vspace{-0.85ex}
\subsection{Objective}
\label{sec:problem-objective}
\vspace{-0.75ex}
The agent's objective is to learn a policy that maps streaming
observations into budgeted retained memory and later answers from that memory:
\[
S_K, w_K \leftarrow \pi(c_{1:K}), \qquad
\hat{y} \sim \pi(\cdot \mid q, S_K).
\]
In the reported retained-memory evaluations, the answer context is the query
plus source excerpts retrieved from $S_K$; the active buffer is fixed across
methods and is not an unbounded raw-log side channel.
Performance is measured by task success, such as answer correctness or F1,
under $B_{\mathrm{ret}}$ and the pre-query protocol.
Section~\ref{sec:method} instantiates this policy with two interfaces: a
pre-query memory writer that builds the retained evidence cover during
ingestion, and a reader that queries this cover after the target query is known.

\vspace{-0.65ex}
\section{Method}
\label{sec:method}
\vspace{-0.65ex}

\subsection{Retention Policy Overview}
\label{sec:retention-policy-overview}
\vspace{-0.7ex}
\name learns a budgeted evidence-retention policy. During ingestion, the writer
decides which source spans should remain in memory under $B_{\mathrm{ret}}$.
The maintained state is a retained evidence cover:
the writer observes the stream online, proposes evidence capsules, and maintains
a cover
$S_t$ under the budget constraint
\[
  |S_t|_{\mathrm{tok}} \le B_{\mathrm{ret}} .
\]
The write path has three parts. Answerability probes identify source spans that
may become useful later. Evidence capsules pair those spans with retrieval keys.
A budget layer admits, consolidates, or rejects proposals under
$B_{\mathrm{ret}}$. At read time, the reader retrieves from this cover and
answers from preserved source evidence.

\vspace{-0.7ex}
\subsection{Answerability Probes}
\label{sec:answerability-probes}
\vspace{-0.7ex}
Topical salience alone is an insufficient write signal: a relevant passage may
omit the fact, relation, date, or update needed later. \name uses answerability
probes, schema-constrained writer decisions that identify what future needs a
chunk may support. A probe specifies which source evidence should remain and how
it should be found; the retained memory stores the source excerpt, not the probe
text.
The policy emits a title, entity anchors, surface and intent retrieval keys, an
update mode, and optional update targets. In extractive mode, the writer selects a retrieval handle, update intent, and bounded source-backed excerpt under budget; the excerpt is later returned verbatim to the reader. Table~\ref{tab:probe-pipeline} summarizes the write path, and Appendix~\ref{app:prompt-templates} gives the schema.

\vspace{-0.7ex}
\begin{table}[H]
  \centering
  \vspace{-0.7ex}
  \caption{Answerability-probe pipeline for one incoming chunk. The policy emits
  retrieval and update fields; retained memory stores source evidence bound
  to those fields.}
  \label{tab:probe-pipeline}
  \scriptsize
  \setlength{\tabcolsep}{3.5pt}
  \resizebox{\linewidth}{!}{%
  \begin{tabular}{@{}lll@{}}
    \toprule
    \textbf{Step} & \textbf{Decision} & \textbf{Materialized field or action} \\
    \midrule
    Evidence test & Keep or skip the chunk & \texttt{update\_mode=insert/merge/overwrite/skip} \\
    Anchor selection & Which entities, dates, or relations identify the fact & \texttt{title}, \texttt{entities} \\
    Retrieval-key generation & How a future query may retrieve the evidence & \texttt{retrieval\_keys\_surface}, \texttt{retrieval\_keys\_intent} \\
    Source binding & Which verbatim source excerpt carries the evidence & bounded \texttt{source\_snippet}/\texttt{focused\_source} \\
    Budget update & Add, consolidate, replace, or reject under budget & \textsc{BudgetUpdate} and update targets \\
    \bottomrule
  \end{tabular}
  }
\end{table}
\vspace{-0.7ex}
When a write action is selected as the training carrier, the GRPO update in
Section~\ref{sec:training} is applied to these emitted memory-control tokens.
The stream text is the writer's input, the future query is hidden during
writing, and the source excerpt is bound from the current chunk.

\vspace{-0.7ex}
\subsection{Source-Backed Evidence Capsules}
\label{sec:evidence-capsules}
\vspace{-0.7ex}
The retained cover consists of evidence capsules: a bounded source excerpt,
retrieval keys, and token cost. The excerpt preserves stream evidence verbatim,
while keys such as entities and intent descriptors make it findable after the query
is known. The main budget counts source-excerpt tokens; retrieval metadata is
reported separately in Appendix Table~\ref{tab:retained-memory-accounting}.
Appendix~\ref{app:memory-writer-prompt} gives the structured capsule action
schema.
The stored object is a grounded capsule, with the source excerpt as the
answer-bearing content.

\vspace{-0.7ex}
\subsection{Budgeted Retention}
\label{sec:budgeted-evidence-cover}
\vspace{-0.7ex}

For each incoming chunk, the writer may propose candidate capsules $M_t$. The
retained cover is updated by a budget-accounting layer:
\[
  S_t \leftarrow \mathrm{BudgetUpdate}(S_{t-1}, M_t; B_{\mathrm{ret}}).
\]
\textsc{BudgetUpdate} is deterministic accounting over learned proposals: it
checks source grounding, counts source-excerpt tokens, and applies valid
updates. During training rollouts, it also records any over-budget mass against
the sampled budget, and the reward penalizes that overrun. Reported evaluations
enforce the fixed retained budget exactly: the memory used at read time satisfies
$|S_K|_{\mathrm{tok}}\le B_{\mathrm{ret}}$. The learned component is the proposal
itself: which spans to preserve, how much source context to include, and which
retrieval keys to attach.
Appendix~\ref{app:prompt-templates} gives the structured action schema, and
Appendix~\ref{app:update-modes} describes the concrete update modes, budget
checks, and consolidation rules used to maintain the cover.

\vspace{-0.7ex}
\subsection{Read-Time Evidence Selection from the Retained Cover}
\label{sec:read-time-selection}
\vspace{-0.7ex}

After the query is known, \name reads only from the retained evidence cover
$S_K$. The reader first expresses its information need as short retrieval
queries $u_{1:n}$:
\[
  C \leftarrow \mathrm{Retrieve}(S_K, u_{1:n}),
\]
where $C$ contains top-$k$ candidate capsules from the retained cover. The
retriever does not search the full raw stream outside $B_{\mathrm{ret}}$; answer
quality therefore depends on evidence retained during ingestion.

The selector then chooses retained evidence for the reader:
\[
  r \sim \pi^{\mathrm{sel}}(\cdot \mid q, C),
\qquad
  \hat{y} \sim \pi^{\mathrm{ans}}(\cdot \mid q, r).
\]
Here $r$ is selected evidence from preserved source excerpts. The bounded active
context is used to manage ingestion and streaming
state, while the reader's answer evidence comes from retained source
excerpts selected out of $S_K$. This read path defines our evidence metrics:
Retain-Recall measures whether gold evidence survives the pre-query budget
update, and Read-Recall measures whether it reaches the reader after the query
is known. Outcome-gated training, described next, aligns both stages with
downstream answer success.

\subsection{Answer-Gated Evidence-Chain Training}
\label{sec:training}

\name trains the memory writer with an answer-gated evidence-chain objective.
The objective rewards retained evidence when it survives ingestion, remains
readable at query time, and supports the final answer under a sampled budget.
This aligns training with the same Survive--Read--Answer bottlenecks measured by
Retain-Recall, Read-Recall, and answer F1. A rollout first commits a retained
evidence cover under
$B_{\mathrm{ret}}$ before the query is known; after the query becomes available,
the reader retrieves from this cover, selects evidence, and answers. The
training signal gives earlier retention decisions their share of the final
answer feedback. Section~\ref{sec:experiments}
describes the training curriculum;
Appendix~\ref{app:training-config} gives the episode construction, budget
sampling, reward instrumentation, and optimizer details.

\paragraph{Objective.}
Each rollout $i$ produces a trajectory $\tau_i$ and an answer $\hat{y}_i$. Let
$Q_i=\mathrm{Verify}(\hat{y}_i,y_i)\in[0,1]$ denote final answer quality. We
implement the chain with a multiplicative answer gate. The survival term is
retained evidence coverage $E_i$; the readability terms are lookup rank score
$L_i$ and selection purity $P_i$; and $W_i$ rewards valid source-preserving
writes. All auxiliary terms are normalized to $[0,1]$. The rollout reward is
\[
\mathcal{R}_i \;=\;
Q_i \,\big(\alpha_Q + \alpha_E E_i + \alpha_L L_i + \alpha_P P_i + \alpha_W W_i\big)
\;-\; \lambda_{\mathrm{budget}}\,
\frac{\max(0, |S_i|_{\mathrm{tok}} - \tilde{B}_i)}{\tilde{B}_i} \,.
\]
Final answer quality gates auxiliary evidence rewards, so survival and
readability proxies receive credit only when they support the eventual answer.
The last term is the budget penalty: if the retained cover $S_i$ exceeds the
sampled budget $\tilde{B}_i$, the overrun is normalized by $\tilde{B}_i$ and
subtracted from the reward. Thus, budget enters the evidence-chain objective:
extra source evidence must improve answer-supported survival or readability
enough to justify its token cost.
This soft penalty is used during training rollouts to give the policy an
explicit budget-aware signal across the frontier; reported evaluations use the
fixed $B_{\mathrm{ret}}$ for comparison.

\paragraph{Group-relative memory-policy update.}
We sample $G=16$ rollouts for each training prompt and compute a single
trajectory reward for each rollout. The reported runs use
$\lambda_{\mathrm{budget}}=0.2$, PPO clip range $\epsilon=0.2$, KL coefficient
$\beta=0.001$, rollout sampling temperature $T=1.0$, a 256-token cap on each
source excerpt, and a 128-token source prefix in the retrieval index. For a
rollout group, we compute the group-relative advantage
\[
\hat{A}_i \;=\; \mathcal{R}_i \;-\; \frac{1}{G}\sum_{j=1}^{G} \mathcal{R}_{j}.
\]
Let $\mathcal{T}_i$ denote the tokens of the selected trainable carrier turn
updated for rollout $i$. In the veRL implementation, the reward is computed from
the full Survive--Read--Answer trajectory, while the PPO/GRPO loss is applied to
one curriculum-selected carrier turn. In write-retention batches, the carrier is
a writer action block emitted during ingestion, such as a capsule proposal or
update action for one chunk window. The rollout then continues through retrieval
and answering, so the selected write decision receives delayed credit from the
final answer-gated reward. Across batches, different write decisions are selected
as carriers, exposing the writer's per-chunk retention decisions to this same
outcome signal. In
read-selection batches, the carrier is the evidence-selection turn; the answer
turn is score-only and does not receive gradient. Appendix~\ref{app:training-config}
gives the token-level clipped objective and implementation details.

\begin{algorithm}[t]
\caption{Answer-Gated Evidence-Chain Training}
\label{alg:grpo}
\begin{algorithmic}[1]
\Require Dataset $\mathcal{D}$, budget grid $\mathcal{G}_{B}=\{512,1024,2048,4096,8192\}$, rollout group size $G=16$
\For{each training batch}
  \For{$i = 1$ to $G$}
    \State Sample episode $(c_{1:K}^{(i)}, q_i, y_i)\sim\mathcal{D}$
    \State Sample budget $\tilde{B}_i \sim \mathrm{Uniform}(\mathcal{G}_{B})$
    \State $(S_i,w_i,\tau_i)\gets\textsc{IngestRetain}(c_{1:K}^{(i)}, \tilde{B}_i)$ \Comment{withhold $q_i$ during ingestion}
    \State $u_i\gets\textsc{Query}(q_i)$; \quad $\mathcal{C}_i\gets\textsc{Retrieve}(S_i,u_i)$
    \State $r_i\gets\textsc{SelectEvidence}(q_i,\mathcal{C}_i)$
    \State $\hat{y}_i\gets\textsc{Answer}(q_i,r_i)$ \Comment{answer from selected retained evidence}
    \State Compute answer-gated evidence-chain reward $\mathcal{R}_i$
    \State Select curriculum carrier $\mathcal{T}_i$ from $\tau_i$ \Comment{write action block or read-selection turn}
  \EndFor
  \State $\bar{\mathcal{R}} \gets \tfrac{1}{G}\sum_{i=1}^{G} \mathcal{R}_i$
  \State $\hat{A}_i \gets \mathcal{R}_i - \bar{\mathcal{R}}$ \quad for each $i \in \{1,\ldots,G\}$
  \State Update only the selected carrier $\mathcal{T}_i$ with a clipped GRPO-style objective
\EndFor
\end{algorithmic}
\end{algorithm}

Algorithm~\ref{alg:grpo} summarizes the rollout and group-relative update.
Appendix~\ref{app:training-config} defines the auxiliary reward terms, episode
format, budget sampling, coefficient selection, token-level objective, and
implementation details.

\vspace{-0.65ex}
\section{Experiments}
\label{sec:experiments}
\vspace{-0.65ex}

\paragraph{Training Details.}
We fine-tune Qwen2.5-7B and Qwen2.5-14B~\citep{yang2024qwen25}
memory-policy backbones on external pre-query episodes, keeping the retriever
fixed during rollouts. Stage I uses
RULER-HotpotQA~\citep{hsieh2024ruler} for controlled multi-hop evidence
preservation and converges in about 500 optimization steps. Stage II then runs a
100-step multi-session continuation from MuSiQue~\citep{trivedi2022musique} and
2WikiMultiHopQA~\citep{ho2020constructing}, with Stage III hard cases for stale
facts, time-scoped evidence, and related-but-unanswerable contexts folded into
the same continuation. Across all stages, the query is hidden during memory
writing, and the budget is sampled from $\{512,1024,2048,4096,8192\}$. The
policy is therefore trained across the full budget frontier.
Checkpoints and reward coefficients are selected on
held-out external pre-query validation episodes, not on LongMemEval-RR or
MultiQ-LongMemEval-RR; the selected coefficient vector
$(\alpha_Q,\alpha_E,\alpha_L,\alpha_P,\alpha_W)=(0.45,0.25,0.15,0.10,0.05)$
is fixed for all reported evaluations.
Appendix~\ref{app:training-config} gives the episode format, optimization
settings, coefficient sensitivity, seed stability, and reward-term ablations.
\vspace{-0.7ex}

\paragraph{Benchmarks.}
LongMemEval-RR~\citep{wu2025longmemeval} is the main external
long-horizon memory evaluation. RULER-HotpotQA~\citep{hsieh2024ruler} is a
controlled, training-adjacent stress test. MultiQ-LongMemEval-RR and ablations
then probe memory reuse and component effects.
\vspace{-0.7ex}
\begin{itemize}[leftmargin=1.5em]
  \item \textbf{LongMemEval-RR} is our LongMemEval-derived Budgeted
  Pre-Query Retention protocol~\citep{wu2025longmemeval}, not an official
  LongMemEval benchmark name. It preserves the original histories, questions,
  and answer targets; only the resource regime changes. Methods ingest sessions
  before the query is known, retain a fixed budget of source evidence, and
  answer from retained memory without returning to the raw stream. This
  evaluates a natural deployment constraint for long-horizon agents: the task
  distribution stays fixed, while query-time raw-log access is no longer
  available. All LongMemEval-derived protocols are used only for evaluation.
  \vspace{-0.7ex}
  \item \textbf{RULER-HotpotQA}~\citep{hsieh2024ruler} is a controlled,
  training-adjacent stress test for evidence survival when memory is written
  before the question. Stage I draws from the same task family, so this benchmark
  probes in-distribution multi-hop retention. LongMemEval-RR provides the
  held-out cross-method comparison.
\end{itemize}
\vspace{-0.7ex}

\paragraph{Baselines.}
We compare three baseline categories: full-log systems with raw-history access,
query-visible memory agents, and budgeted evidence-survival methods. \name
belongs to the last category: it writes memory before the query is known and
answers only from retained memory. On RULER-HotpotQA, we compare with
context-only Qwen models, fixed-retriever vanilla RAG, and a pre-query
adaptation of MemAgent; Appendix~\ref{app:memagent-prequery} reports its
adaptation curve. Because MemAgent was designed for query-visible in-context
memory rewriting, this adapted version serves as a diagnostic, protocol-matched
baseline for source-evidence retention.
\vspace{-0.7ex}

\paragraph{LongMemEval-RR Main Results.}
Table~\ref{tab:longmemeval-rr-main} evaluates the evidence-survival chain at
$B_{\mathrm{ret}}=8192$, the largest budget used by the budgeted baselines. We
use this common max-budget anchor to give every baseline its largest source
allowance. At this anchor, \name-14B reaches 0.3017 F1,
compared with the top non-\name budgeted F1 baseline at 0.1765, a +0.125 F1 gain
(95\% paired bootstrap CI: [+0.081,+0.169]).
The closest baseline is TierMem-BudgetRaw, whose F1 is competitive despite low
Retain-Recall and Read-Recall because compressed summaries can still carry
answer signal without appearing as source-traceable gold evidence. \name's
advantage is that the answer gain travels through the source-traceable evidence
chain: more gold evidence survives, more of it is read back, and final F1 rises.
Figure~\ref{fig:budget-frontier} gives the corresponding efficiency frontier:
at every budget level, \name converts the same $B_{\mathrm{ret}}$ into higher
F1, Retain-Recall, and Read-Recall than the budgeted baselines. The tight-budget
setting is already effective: at $B_{\mathrm{ret}}=512$, \name-7B reaches
0.2413 F1, above the strongest non-\name budgeted
baseline at $B_{\mathrm{ret}}=8192$. This uses one-sixteenth of the budget,
keeping about 7.7K fewer source tokens in the query-time memory state. For
long-horizon agents, this difference becomes query-time cost: if retained memory
misses useful evidence, the system must retrieve or reread larger raw contexts.
At this tightest budget, \name-7B has slightly higher F1 than \name-14B despite
lower Retain-Recall and Read-Recall, a small reader-side crossover within the
same evidence-survival frontier.
This pattern matches the training design, where the writer samples
multiple budgets and learns to choose evidence under both tight and loose memory
limits. Oracle retention reaches 0.4499 F1 and 0.8367 Read-Recall at the same
$8192$-token anchor. The full
sweep in Appendix~Table~\ref{tab:longmemeval-rr-f1-budget-sweep} shows how the
remaining gap can move to the reader side: as the retained cover grows, fixed
top-$k$ retrieval can introduce candidate competition, so Read-Recall and F1
need not rise monotonically even when gold evidence survives ingestion.
Tables~\ref{tab:retained-memory-accounting} and
\ref{tab:serialized-budget-check} make the metadata accounting explicit,
including a conservative serialized-budget check.
\vspace{-0.7ex}

\paragraph{Mechanism Ablation: Survive, Read, Answer.}
Table~\ref{tab:longmemeval-rr-ablation-main} follows evidence through
the pipeline: survival after ingestion, access at read time, and final answer
F1. The ablations show that source evidence, answerability probes, and retrieval
keys are coupled. Without probes, the writer keeps weaker source evidence:
Retain-Recall falls from 0.3215 to 0.2426. Without retrieval keys, source
survival is largely preserved, but read-time access drops from 0.3112 to 0.2162:
the evidence is retained yet hard to recover. Full \name resolves both failures:
probes choose source spans, retrieval keys bring them back at read time, and
delayed outcome training aligns the retained cover with final answer quality.
The row labeled
\name w/o outcome RL uses the same
probe-and-capsule interface as \name; its gap to full \name isolates the effect
of answer-gated training on write-time retention choices. The reader and prompt
format stay fixed.
\vspace{-0.7ex}

\begin{table}[H]
  \centering
  \caption{LongMemEval-RR mechanism ablations at $B=8192$. Columns track
  retention after ingestion, read-time access, and answer F1.}
  \label{tab:longmemeval-rr-ablation-main}
  
  \scriptsize
  \setlength{\tabcolsep}{3.5pt}
  \resizebox{\linewidth}{!}{%
  \begin{tabular}{@{}llcccl@{}}
    \toprule
    \textbf{Variant} & \textbf{Mechanism removed} &
    \textbf{Retain-Recall} & \textbf{Read-Recall} &
    \textbf{F1} & \textbf{Failure mode} \\
    \midrule
    \name full & -- & 0.3215 & 0.3112 & 0.3017 & Missed evidence / Aggregation errors \\
    \name w/o outcome RL & Delayed outcome training & 0.2651 & 0.2481 & 0.2242 & Weak outcome alignment \\
    No answerability probes & Write-time answerability cues & 0.2426 & 0.2075 & 0.2031 & Poor source selection \\
    No retrieval keys & Retrieval keys & 0.3032 & 0.2162 & 0.2089 & Retained source is hard to retrieve/use \\
    Generated-query indexing~\citep{nogueira2019doc2query} & Answerability-grounded cues & 0.2021 & 0.1423 & 0.1205 & Query-like keys do not preserve answerability \\
    Summary-only & Source traceability & N/A & N/A & 0.1156 & Compressed memory loses evidence \\
    Oracle retention & Learned retention & 0.9998 & 0.8367 & 0.4499 & $B=8192$ oracle reference \\
    \bottomrule
  \end{tabular}
  }
  
\end{table}
\vspace{-0.7ex}

\paragraph{MultiQ Coverage Ablation.}
MultiQ-LongMemEval-RR tests whether one frozen retained memory can
serve multiple future tasks. Table~\ref{tab:multiq-longmemeval-rr-main} shows
the failure mode of generic budgeted retention: strong heuristics preserve only
10--12\% of future evidence at a 10\% budget and leave coverage balance at zero.
\name-14B preserves a broader cover and reaches 0.2767 F1, compared with 0.0824
for the strongest heuristic baseline.
\vspace{-0.7ex}

\begin{table}[H]
  \centering
  \caption{MultiQ-LongMemEval-RR retention. One retained
  memory serves multiple future queries over the same history. Values report
  the 10\% budget with top-$k=10$
  retrieval. Coverage balance is the average weakest-query
  Retain-Recall.}
  \label{tab:multiq-longmemeval-rr-main}
  
  \scriptsize
  \setlength{\tabcolsep}{4pt}
  \resizebox{\linewidth}{!}{%
  \begin{tabular}{@{}lcccccc@{}}
    \toprule
    \textbf{Method} & \textbf{Budget} & \textbf{Mean-query Retain-Recall} &
    \textbf{Read-Recall} & \textbf{Group Retain-Recall} &
    \textbf{Coverage balance} & \textbf{F1} \\
    \midrule
    Recency-$B$ & 10\% & 0.1200 & 0.1200 & 0.1213 & 0.0000 & 0.0451 \\
    Reservoir-$B$ & 10\% & 0.1157 & 0.1157 & 0.1082 & 0.0000 & 0.0483 \\
    Dense MMR & 10\% & 0.1131 & 0.1131 & 0.1141 & 0.0000 & 0.0614 \\
    Hybrid salience & 10\% & 0.1063 & 0.1063 & 0.1071 & 0.0000 & 0.0824 \\
    Oracle retention & 10\% & 0.6275 & 0.6275 & 0.6083 & 0.0517 & 0.3982 \\
    \midrule
    \name-7B & 10\% & 0.2752 & 0.2691 & 0.3218 & 0.0312 & 0.2617 \\
    \name-14B & 10\% & 0.3081 & 0.3051 & 0.3412 & 0.0356 & 0.2767 \\
    \bottomrule
  \end{tabular}
  }
 \vspace{0.35ex}
\end{table}

\begin{figure}[H]
  \centering
  
  \includegraphics[width=0.66\linewidth]{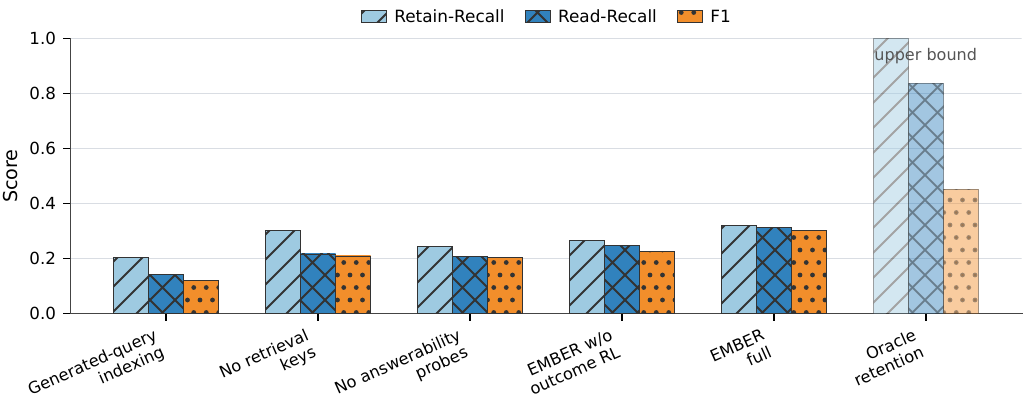}
  
  \caption{Evidence-loss chain on LongMemEval-RR. Each variant is evaluated at
  $B=8192$ with the same reader. Bars show where evidence is lost: ingestion,
  read-time access, or answer use.}
  \label{fig:evidence-loss-chain}
\end{figure}
\vspace{-0.7ex}

\paragraph{Controlled Pre-Query QA.}
RULER-HotpotQA is an in-distribution stress test for \name, not the paper's
held-out benchmark. Stage I trains on the same task family, so the question is
deliberately narrow: whether a writer blind to the query can preserve a clean
multi-hop support chain through the retained-budget bottleneck. At the 28K
operating point,
Table~\ref{tab:ruler-hotpotqa-main} shows \name-14B at $0.8412{\pm}0.033$ F1,
above the strongest vanilla RAG configuration at $0.7772{\pm}0.037$ F1.
\name also exceeds the query-visible references, but those rows run different
protocols and are not matched to \name's training distribution. We therefore use
RULER-HotpotQA as a controlled stress test; LongMemEval-RR remains the held-out
cross-method comparison. Appendix~\ref{app:prequery-length-sweep} reports the
full length sweep.
\vspace{-0.7ex}

\begin{table}[H]
  \centering
  \caption{Controlled, training-adjacent pre-query QA on
  RULER-HotpotQA at 28K tokens. Appendix~\ref{app:prequery-length-sweep} reports
  the full length sweep. F1 is reported as mean $\pm$ 95\% confidence-interval
  half-width.}
  \label{tab:ruler-hotpotqa-main}
  \scriptsize
  \setlength{\tabcolsep}{3.5pt}
  \resizebox{\linewidth}{!}{%
  \begin{tabular}{@{}llllr@{}}
    \toprule
    \textbf{Group} & \textbf{Method} &
    \textbf{Query visible at write time?} &
    \textbf{Full source at read time?} &
    \textbf{F1 $\uparrow$} \\
    \midrule
    Context-only & Qwen3-8B & No & Yes & $0.7108{\pm}0.041$ \\
    Full-source retrieval & Vanilla RAG, Qwen3-8B, $k{=}8$ & No & Yes & $0.7772{\pm}0.037$ \\
    Query-visible reference & MemAgent-7B~\citep{yu2026memagent} & Yes & No & $0.7865{\pm}0.037$ \\
    Query-visible reference & Memory-R1~\citep{yan2025memoryr1} & Yes & No & $0.8165{\pm}0.035$ \\
    \midrule
    Budgeted evidence survival & MemAgent-7B (pre-query adapted)~\citep{yu2026memagent} & No & No & $0.2606{\pm}0.039$ \\
    Budgeted evidence survival & Memory-R1 (pre-query adapted)~\citep{yan2025memoryr1} & No & No & $0.3124{\pm}0.041$ \\
    Budgeted evidence survival & Source-snippet retention & No & No & $0.7353{\pm}0.039$ \\
    \midrule
    Budgeted evidence survival & \textbf{\name-7B} (Qwen2.5-7B) & No & No & $\mathbf{0.8195{\pm}0.034}$ \\
    Budgeted evidence survival & \textbf{\name-14B} (Qwen2.5-14B) & No & No & $\mathbf{0.8412{\pm}0.033}$ \\
    \midrule
    \multicolumn{4}{@{}l}{$\Delta$ vs strongest Vanilla RAG} & \textbf{+0.0640} \\
    \bottomrule
  \end{tabular}
  }
  \vspace{0.35ex}
  \footnotesize
	  \parbox{0.98\linewidth}{\textit{Notes.} Budgeted evidence-survival rows write
	  memory before the question is known and answer without full-source
	  access. RULER-HotpotQA is in the Stage-I task family, so this table is a
	  controlled in-distribution stress test; the held-out cross-method comparison
	  is LongMemEval-RR. Full-source and query-visible rows are reference
	  categories, not distribution-matched head-to-head baselines.}
\end{table}
\vspace{-0.7ex}

\vspace{-0.65ex}
\section{Related Work}
\label{sec:related}
\vspace{-0.65ex}

\paragraph{Agent memory and learned memory operations.}
Persistent-memory systems store long-lived interaction records or
external memory substrates for later use~\citep{park2023generative,zhong2023memorybank,
packer2023memgpt,wang2023longmem,wang2024memoryllm,chhikara2025mem0,arigraph2024,amem2025}.
Recent learned memory-operation methods train agents to rewrite, update,
revisit, or use memory entries~\citep{yu2026memagent,yan2025memoryr1,
wang2025memalpha,huo2026atommem,shi2026rememr1}. MemAgent and Memory-R1 are the
closest examples: they make memory writing an explicit learned action over
entries or rewrites. The retention question remains: before the query arrives,
which source evidence should fit inside $B_{\mathrm{ret}}$ and remain useful
after ingestion?

\vspace{-0.65ex}
\paragraph{RAG, tiered memory, and budgeted retrieval.}
Retrieval-augmented and tiered-memory systems reduce read-time cost by
indexing, routing, or compressing access to stored information~\citep{lewis2020retrieval,
zhang2026budgetmem,fang2025lightmem,zhu2026tiermem}. TierMem is the sharpest
contrast: it routes among summaries and can escalate to an immutable raw-log store
when summary evidence is insufficient. In our setting, the full-log fallback is
outside the query-time budget, so the retention decision must happen during
ingestion: which source evidence enters the small retained memory, and will it
still support later answers?

\vspace{-0.65ex}
\paragraph{Query prediction, summarization, and context compression.}
Document expansion, self-questioning, and evidence-utility methods
improve retrieval or memory access through predicted queries, self-generated
probes, or utility signals~\citep{nogueira2019doc2query,yang2026promem,
ma2025nemori,jain2026cuer}. Extractive and query-focused
summarization select source spans under a length budget or for a known
information need~\citep{liu2019bertsum,dong2018banditsum,dang2006duc}: the
objective is document coverage or relevance to an observed query. Related
context- and cache-compression work reduces read-time text or hidden-state
footprint. These directions reduce the cost of using an available context or
corpus. \name stores source excerpts with write-time answerability cues and
retrieval keys, so retained evidence can be recovered under the query-time
budget.

\vspace{-0.75ex}
\section{Conclusion}
\label{sec:conclusion}
\vspace{-0.75ex}
This paper treats long-horizon memory as a retained evidence state under a fixed
token budget. The results on LongMemEval-RR and MultiQ-LongMemEval-RR show that
a compact memory can outperform generic retention and budgeted raw fallback when
the write step preserves evidence the reader can later recover. This view makes
memory quality auditable: the retained state can be inspected for the source
evidence it carries, the evidence it loses, and the retrieval failures that
remain. Appendix~\ref{app:retained-evidence-state} discusses how the same
evidence-state view extends to persistent agent memory.

\vspace{-0.75ex}
\section{Limitations and Broader Impact}
\label{sec:limitations-impact}
\vspace{-0.75ex}
We evaluate \name on controlled long-horizon memory benchmarks, not deployed
assistant systems. Scores depend on evidence-label granularity, retained-token
budget, and the reader/retriever; deployments may add variation from user-specific
histories, evolving preferences, privacy constraints, and memory drift. We estimate
training-seed variance with three \name-7B runs.
Selective retention can reduce the amount of history read at query time, making
memory more compact and auditable, but it can also preserve sensitive, stale, or
misleading evidence. Deployed systems should provide inspection and deletion controls, respect
user intent before preserving private evidence, and treat retained memory as
auditable application state, not invisible model context.

\nocite{yu2026memagent,guo2025deepseekr1,jin2025searchr1,chhikara2025mem0}
\bibliographystyle{unsrtnat}
\bibliography{references}

\clearpage
\appendix
\section{Prompt Templates (Schema-Constrained JSON)}
\label{app:prompt-templates}

\subsection{Memory Writer Prompt}
\label{app:memory-writer-prompt}

\begin{PromptBlock}
SYSTEM: You are a memory writer. Output ONLY valid JSON.
GOAL: Convert the current stream chunk into source-evidence capsule actions
and a compact residual context.
CONSTRAINTS:
- produce source-evidence capsule actions when the chunk contains durable evidence
- emit a skip action when the chunk adds no evidence worth retaining
- each action uses insert / merge / overwrite / skip
- keep retrieval keys concise
- preserve source evidence needed for future answers through source excerpts
- residual active context must be compact
INPUTS:
- current_chunk: ...
- candidate_context: ...
- related_capsules:
  [ {id, title, entities, retrieval_keys, source_excerpt, version, update_mode}, ... ]
- retained_memory_state: ...
- usage_signals: ...
- budget_state: ...
OUTPUT JSON SCHEMA:
{
  "memory_items": [
    {
      "title": "...",
      "entities": ["...", "..."],
      "retrieval_keys_surface": ["...", "..."],
      "retrieval_keys_intent": ["...", "..."],
      "focused_source": "...",
      "update_mode": "insert|merge|overwrite|skip",
      "merge_target_id": "..."
    }
  ],
  "residual_context": {
    "pointers": ["...", "..."],
    "active_frontier": ["...", "..."],
    "residual_local_context": ["...", "..."]
  }
}
NOTE: This JSON is the memory-writer action format. The stored form is a
retrieval-indexed source-evidence capsule materialized from these fields.
For \texttt{skip}, the writer emits only the update mode and no source payload.
\end{PromptBlock}

\subsection{Task-Solver Prompt (Query Generation)}
\label{app:task-solver-query-prompt}

\begin{PromptBlock}
SYSTEM: You are a retrieval query generator. Output ONLY JSON.
GOAL: Produce 1-3 short retrieval queries for the current question.
CONSTRAINTS:
- each query <= 12 tokens
- diversify queries: entity-based + intent-based
OUTPUT:
{"queries": ["...", "..."]}
\end{PromptBlock}

\clearpage
\subsection{Memory Writer Evidence Selection Prompt}
\label{app:memory-writer-evidence-selection-prompt}

\begin{PromptBlock}
SYSTEM: You are a memory writer selecting retained source evidence.
Output ONLY JSON.
GOAL: From retrieved memory items, select the preserved source evidence
that should return to the task solver.
CONSTRAINTS:
- select the most relevant item ids for the given query
- avoid redundant or weakly supported items
- do not rewrite or compress the source excerpts
- the answer prompt will receive the selected source excerpts verbatim
INPUTS:
- query: ...
- active_context: ...
- retrieved_candidates: [ {id, title, entities, retrieval_keys, source_excerpt}, ... ]
OUTPUT JSON SCHEMA:
{
  "selected_ids": ["id1", "id2"]
}
NOTE: The task solver sees the source excerpts attached to the selected
items. The memory writer uses the retained-memory structure to judge
which retrieved evidence is authoritative and non-redundant.
\end{PromptBlock}

\section{Implementation Notes}
\label{app:implementation-notes}

\subsection{Update Modes for the Retained Evidence Cover}
\label{app:update-modes}

The writer proposes each capsule with an \texttt{update\_mode} field. The
budget-accounting layer executes the proposal deterministically:

\begin{description}[leftmargin=1.5em,style=nextline]
  \item[\texttt{insert}] Adds a new capsule to the cover. The layer builds a
    stored item from the proposed source excerpt, anchors, and retrieval keys,
    indexes it, and charges its source-evidence token cost against~$B_{\mathrm{ret}}$.
    If admission would exceed the budget, the proposal is rejected.
  \item[\texttt{merge}] Updates an existing capsule identified by
    \texttt{merge\_target\_id}. New facts are appended and deduplicated; the
    title, entities, and retrieval keys are replaced with the proposal's values.
    The source excerpt is overwritten only when the proposal supplies a new one;
    otherwise the original excerpt is preserved. The merged capsule is
    re-embedded and re-indexed.
  \item[\texttt{overwrite}] Replaces the target capsule entirely: all fields
    are set to the proposed values, the version counter increments, and the
    index is rebuilt. This mode is used when the writer determines that the
    existing capsule is stale or superseded.
  \item[\texttt{skip}] No modification to the cover. The writer emits
    \texttt{skip} when the current observation adds no evidence worth retaining.
    The layer validates that no target ID or payload accompanies a skip action.
\end{description}

\paragraph{Budget rejection.}
Every non-skip proposal is charged a source-evidence token cost equal to the length of
its source excerpt. If the cumulative retained tokens after the proposed update
would exceed~$B_{\mathrm{ret}}$, the proposal is rejected and the cover remains
unchanged. During training rollouts, over-budget proposals incur a cost penalty
in the reward; during evaluation, the budget is enforced exactly.

\paragraph{Deduplication and consolidation.}
Before insertion, the layer checks for near-duplicate capsules using embedding
cosine similarity. When a proposed capsule overlaps substantially with an
existing one, the writer is expected to use \texttt{merge} or
\texttt{overwrite} rather than \texttt{insert}. Invalid references (e.g., a
\texttt{merge} targeting a nonexistent ID) trigger a validation error and
the proposal is dropped.

\subsection{General Implementation Practices}
\label{app:general-impl}

\begin{itemize}[leftmargin=1.5em]
  \item Strict JSON validation on every writer output; invalid outputs are
    resampled and resamples are counted as cost when desired.
  \item A lightweight dedup index (embeddings + hashing) identifies
    near-duplicate candidates for the merge and overwrite modes above.
  \item The retriever remains frozen during RL rollouts for training stability
    and a fixed read-side interface.
\end{itemize}

\section{\texorpdfstring{Appendix}{Appendix}}
\label{app:appendix}

\subsection{\texorpdfstring{Retained Evidence State}{Retained Evidence State}}
\label{app:retained-evidence-state}

Long-horizon memory is more than an index, a summary, or an extended context; it
is a retained evidence state built before the next request arrives. Retain-Recall
and Read-Recall make this state measurable by separating write-time evidence
loss from read-time access failure and answer-time reasoning failure. This
decomposition lets Budgeted Pre-Query Retention test memory construction rather
than only retrieval quality.

Persistent assistants, coding agents, and tool-using systems accumulate
dialogue, files, edits, and observations before the next request arrives. \name
does not require these systems to archive everything as raw context; it provides
a learnable interface for deciding which source evidence should remain available
under $B_{\mathrm{ret}}$.

\subsection{\texorpdfstring{Vanilla RAG K-Curve Control}{Vanilla RAG K-Curve Control}}
\label{app:k-curve}

\begin{table}[htb]
  \centering
  \caption{Vanilla RAG K-curve on the 28K-token pre-query
  RULER-HotpotQA setting, separate from the LongMemEval-RR retained-budget
  protocol. All rows use the Qwen2.5-7B backbone and BGE-small~\citep{bge_embedding}
  retriever. The control shows that \name's extractive read-select gain is not
  explained by reduced context size alone. Values report answer F1 in percentage
  points.}
  \label{tab:k-curve}
  
  \small
  \setlength{\tabcolsep}{5pt}
  \begin{tabular}{@{}lc@{}}
    \toprule
    \textbf{Method} & \textbf{Result} \\
    \midrule
    Qwen2.5-7B + vanilla RAG ($k=3$) & 62.28 \\
    Qwen2.5-7B + vanilla RAG ($k=8$) & 73.08 \\
    Qwen2.5-7B + vanilla RAG ($k=15$) & 74.84 \\
    Qwen2.5-7B + extractive read-select & 76.95 \\
    \bottomrule
  \end{tabular}
  
\end{table}

\subsection{\texorpdfstring{MemAgent Retraining for Pre-Query Writing}{MemAgent Retraining for Pre-Query Writing}}
\label{app:memagent-prequery}

For the MemAgent baseline, we follow the MemAgent in-context memory
algorithm but train and evaluate it under the same pre-query protocol used
for \name. During memory-writing turns, the model receives the stream chunk and
its current in-context memory state, but not the downstream question. The
question becomes available only after memory construction, when the model
retrieves or uses the maintained memory to answer. This is the pre-query adapted
MemAgent baseline used in the main comparison. The adaptation
tests whether MemAgent's in-context rewriting interface can operate under the
same timing constraint; it is not the native query-visible MemAgent setting. In
the held-out curve below, retraining raises validation F1 from 0.0180 to 0.1330
and lowers unknown answers from 0.9860 to 0.8140, but the adapted interface
remains weaker than source-evidence retention methods in the main comparison.

\begin{figure}[H]
  \centering
  
  \includegraphics[width=0.74\linewidth]{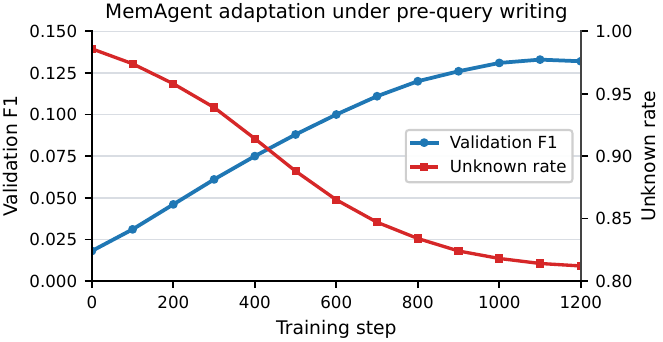}
  
  \caption{Retraining curve for MemAgent adapted to pre-query
  writing. The curve tracks the held-out pre-query objective during adaptation:
  memory is written before the target question is known, then evaluated after
  the question becomes available.}
  \label{fig:memagent-adapted-retrain-curve}
\end{figure}
\vspace{-0.7ex}

\subsection{\texorpdfstring{LongMemEval-RR Protocol and Extended Tables}{LongMemEval-RR Protocol and Extended Tables}}
\label{app:longmemeval-rr}

LongMemEval-RR is a derived evaluation protocol built on
LongMemEval~\citep{wu2025longmemeval}. It reuses LongMemEval histories and
questions, keeps the answer targets fixed, and changes only access timing and
the retained evidence budget. Methods ingest sessions online before
the target query is known; after ingestion, they may use only a fixed budget
  $B_{\mathrm{ret}}$ of retained source evidence. The full raw history is outside
  this budget at query time, and answers must be produced from retained memory. We
  report Retain-Recall, Read-Recall, answer F1, Sub-EM, and budget-normalized
  metrics. The offline retention probe measures whether gold evidence
  survives $B_{\mathrm{ret}}$. Table~\ref{tab:longmemeval-rr-reader-check}
  reports GPT-4o reader evaluations after query-time retrieval.
  We release the construction script, budget settings, chunk/span rules,
  retained-memory wrapper, and evaluator with the benchmark artifacts.

\begin{table}[H]
  \centering
  \caption{LongMemEval-RR protocol card. LongMemEval-RR is a derived retained-memory
  protocol, not an official LongMemEval benchmark name. We identify the exact
  input snapshot by file hash for reproducibility.}
  \label{tab:longmemeval-rr-protocol-card}
  \scriptsize
  \setlength{\tabcolsep}{4pt}
  \resizebox{\textwidth}{!}{%
  \begin{tabular}{@{}lp{0.72\linewidth}@{}}
    \toprule
    \textbf{Item} & \textbf{Protocol specification} \\
    \midrule
    Source snapshot &
    LongMemEval-S JSON from \texttt{xiaowu0162/longmemeval}; input file
    \texttt{longmemeval\_s.json}, SHA-256
    \texttt{dad95067c042691bd270bb304f02c5dad46c434874346d819181b827d5b88055}. \\
    Examples used &
    All 500 examples in the snapshot are used for the main LongMemEval-RR
    reader evaluation; no main-result examples are dropped. \\
    Source-evidence units &
    Main tables use turn-level units. Each turn is rendered with session ID,
    date, speaker role, and turn text. Session-level probes are reported
    separately as diagnostics. \\
    Gold evidence labels &
    Session-level gold units come from the native \texttt{answer\_session\_ids}
    field. Turn-level gold units use native turn \texttt{has\_answer} labels
    when present; if no positive turn label is present for an example, all turns
    in the annotated answer sessions are treated as gold. We do not construct
    LongMemEval-RR gold labels by answer-string matching. \\
    Oracle retention &
    Oracle retention packs these gold source-evidence units into the same
    $B_{\mathrm{ret}}$. It uses gold evidence labels during
    retention but receives no answer text at read time. \\
    MultiQ construction &
    MultiQ-LongMemEval-RR groups five future questions over one compiled
    history. We construct 100 groups with seed 42 and target 80 sessions per
    group; question text is stored only as post-ingest evaluation metadata, not
    as an ingest-time session. \\
    Scripts &
    {\raggedright Loader: \texttt{longmemeval.py}. Retention metrics:
    \texttt{eval\_retention\_probe.py}. Reader evaluation:
    \texttt{eval\_longmemeval\_rr\_reader.py}. MultiQ builder:
    \texttt{build\_multiq\_longmemeval\_rr.py}.} \\
    \bottomrule
  \end{tabular}
  }
\end{table}

\paragraph{\texorpdfstring{Retention metrics.}{Retention metrics.}}
For each query $q$, let $G(q)$ denote the gold source-evidence
units, $M(q)$ the source-evidence units retained after ingestion, and
$R(q)$ the units retrieved for the reader after the query becomes available. We
measure the two points at which answer evidence can be lost:
\[
\operatorname{RetainRecall}(q)=\frac{|G(q)\cap M(q)|}{|G(q)|},
\qquad
\operatorname{ReadRecall}(q)=\frac{|G(q)\cap R(q)|}{|G(q)|}.
\]
Retain-Recall measures retention: how much gold evidence remains in
memory after pre-query ingestion. Read-Recall measures answer-time access:
how much gold evidence reaches the reader context after query-time retrieval.
For budgeted methods, retrieval is performed from retained memory, so
Read-Recall is upper-bounded by Retain-Recall. Low Retain-Recall identifies a
write-time retention failure; high Retain-Recall with low Read-Recall identifies
a retrieval or ranking failure; high Read-Recall with low F1 identifies a
reader-side reasoning or aggregation failure. Methods that do not retain
source-traceable units, such as abstractive summaries or native MemAgent memory
entries, report N/A for these source-evidence metrics. Summary-derived signals
can still help answer F1, but they are not counted as retained gold evidence
unless they preserve source-traceable units. This distinction explains why a
summary-heavy baseline can obtain competitive F1 while showing low
Retain-Recall and Read-Recall.
\paragraph{\texorpdfstring{Statistical reporting.}{Statistical reporting.}}
Main reader tables define their uncertainty notation in the corresponding
captions. LongMemEval-RR reports bootstrap confidence half-widths; the
RULER-HotpotQA headline table reports 95\% confidence-interval half-widths. Extended
deterministic diagnostics report point estimates.
\vspace{-0.7ex}

\begin{table}[H]
  \centering
  \caption{Retention sweep and fixed-budget evidence results on
  LongMemEval-RR (turn-level, top-$k=10$). Values are reported across
  $B_{\mathrm{ret}}$.}
  \label{tab:longmemeval-rr-budget-sweep}
  
  \scriptsize
  \setlength{\tabcolsep}{3pt}
  \resizebox{\textwidth}{!}{%
  \begin{tabular}{@{}llccccc@{}}
    \toprule
    \textbf{Method} & \textbf{Metric} & \textbf{$B{=}512$} & \textbf{$B{=}1024$} & \textbf{$B{=}2048$} & \textbf{$B{=}4096$} & \textbf{$B{=}8192$} \\
    \midrule
    Recency-$B$ & Retain-Recall & 0.0096 & 0.0202 & 0.0380 & 0.0670 & 0.1521 \\
    Recency-$B$ & Read-Recall & 0.0096 & 0.0202 & 0.0335 & 0.0607 & 0.0943 \\
    Dense MMR & Retain-Recall & 0.0180 & 0.0409 & 0.0754 & 0.1210 & 0.2161 \\
    Dense MMR & Read-Recall & 0.0180 & 0.0407 & 0.0709 & 0.1151 & 0.1912 \\
    Hybrid salience & Retain-Recall & 0.0241 & 0.0485 & 0.0852 & 0.1460 & 0.2140 \\
    Hybrid salience & Read-Recall & 0.0213 & 0.0421 & 0.0752 & 0.1270 & 0.1763 \\
    Source-snippet heuristic & Retain-Recall & 0.0256 & 0.0712 & 0.1141 & 0.1541 & 0.2657 \\
    Source-snippet heuristic & Read-Recall & 0.0231 & 0.0658 & 0.0812 & 0.1258 & 0.1754 \\
    Oracle retention & Retain-Recall & 0.9608 & 0.9744 & 0.9839 & 0.9960 & 0.9998 \\
    Oracle retention & Read-Recall & 0.9576 & 0.9602 & 0.9219 & 0.8842 & 0.8367 \\
    \name-7B & Retain-Recall & 0.1116 & 0.1295 & 0.1845 & 0.2343 & 0.2966 \\
    \name-7B & Read-Recall & 0.0987 & 0.1017 & 0.1756 & 0.2116 & 0.2915 \\
    \name-14B & Retain-Recall & 0.1203 & 0.1385 & 0.2016 & 0.2543 & 0.3215 \\
    \name-14B & Read-Recall & 0.1095 & 0.1167 & 0.1821 & 0.2345 & 0.3112 \\
    \bottomrule
  \end{tabular}
  }
  
\end{table}

\begin{table}[H]
  \centering
  \caption{Memory-token efficiency on LongMemEval-RR (turn-level, GPT-4o
  reader, top-$k=10$). Table~\ref{tab:longmemeval-rr-main} uses $B=8192$ as the
  fixed cross-method comparison point. F1 per 1K retained source-evidence tokens is computed only for
  budgeted methods with source-traceable retained evidence. Full-log reference
  systems are excluded because their accessible raw-log denominator is not the
  retained evidence budget.}
  \label{tab:longmemeval-rr-f1-per-token}
  
  \scriptsize
  \setlength{\tabcolsep}{4pt}
  \resizebox{\textwidth}{!}{%
  \begin{tabular}{@{}lccccc@{}}
    \toprule
    \textbf{Method} & \textbf{Retained source-evidence tokens} &
    \textbf{Retain-Recall} & \textbf{Read-Recall} &
    \textbf{F1} & \textbf{F1 / 1K retained source-evidence tokens} \\
    \midrule
    Random-$B$ & 8192 & 0.0612 & 0.0425 & 0.0350 & 0.0043 \\
    Recency-$B$ & 8192 & 0.1521 & 0.0943 & 0.0827 & 0.0101 \\
    Reservoir-$B$ & 8192 & 0.0753 & 0.0615 & 0.0498 & 0.0061 \\
    Hybrid salience & 8192 & 0.2140 & 0.1763 & 0.1538 & 0.0188 \\
    Source-snippet heuristic & 8192 & 0.2657 & 0.1754 & 0.1763 & 0.0215 \\
    Generated-query indexing~\citep{nogueira2019doc2query} & 8192 & 0.2021 & 0.1423 & 0.1205 & 0.0147 \\
    TierMem-BudgetRaw~\citep{zhu2026tiermem} & 8192 & 0.1246 & 0.0997 & 0.1765 & 0.0215 \\
    Memory-R1 (pre-query adapted)~\citep{yan2025memoryr1} & 8192 & 0.1714 & 0.1464 & 0.1565 & 0.0191 \\
    Oracle retention & 8192 & 0.9998 & 0.8367 & 0.4499 & 0.0549 \\
    \midrule
    \name-7B & 8192 & 0.2966 & 0.2915 & 0.2768 & 0.0338 \\
    \name-14B & 8192 & 0.3215 & 0.3112 & 0.3017 & 0.0368 \\
    \bottomrule
  \end{tabular}
  }
  
\end{table}

\begin{table}[H]
  \centering
  \caption{Retained-memory accounting at the $B=8192$ LongMemEval-RR operating
  point. All rows use the same 8192-token retained source-evidence cap. The main
  protocol charges
  source-excerpt tokens because they are the answer-bearing
  content. Retrieval metadata is serialized as index text and reported
  separately; it is not returned as answer content, and the reader receives
  source excerpts selected through the index. For \name, this overhead is at
  most 18.5\% of the source-evidence cap.}
  \label{tab:retained-memory-accounting}

  \scriptsize
  \setlength{\tabcolsep}{3pt}
  \resizebox{\textwidth}{!}{%
  \begin{tabular}{@{}lrrrrr@{}}
    \toprule
    \textbf{Method} &
    \textbf{Source-evidence cap} &
    \textbf{Avg metadata tokens} &
    \textbf{Total serialized tokens} &
    \textbf{Metadata/source ratio} &
    \textbf{F1} \\
    \midrule
    Source-snippet heuristic & 8192 & 0 & 8192 & 0.0\% & 0.1763 \\
    Generated-query indexing over snippets~\citep{nogueira2019doc2query} & 8192 & 1657 & 9849 & 20.2\% & 0.1205 \\
    \name-7B & 8192 & 1243 & 9435 & 15.2\% & 0.2768 \\
    \name-14B & 8192 & 1516 & 9708 & 18.5\% & 0.3017 \\
    \bottomrule
  \end{tabular}
  }
\end{table}

\begin{table}[H]
  \centering
  \caption{Serialized-budget accounting check on LongMemEval-RR. The main
  protocol budgets retained source-evidence tokens and reports metadata
  separately in Table~\ref{tab:retained-memory-accounting}. This stricter check
  charges serialized index text against an 8192-token serialized-token envelope. The
  \name rows use the existing $B=4096$ reader runs and charge the larger
  $B=8192$ metadata overhead from Table~\ref{tab:retained-memory-accounting} as
  a conservative upper bound; their total serialized footprints remain below
  8192 tokens.}
  \label{tab:serialized-budget-check}

  \scriptsize
  \setlength{\tabcolsep}{3pt}
  \resizebox{\textwidth}{!}{%
  \begin{tabular}{@{}lrrrr@{}}
    \toprule
    \textbf{Method} &
    \textbf{Source-evidence cap} &
    \textbf{Metadata charge} &
    \textbf{Serialized footprint for check} &
    \textbf{F1} \\
    \midrule
    Source-snippet heuristic & 8192 & 0 & 8192 & 0.1763 \\
    TierMem-BudgetRaw~\citep{zhu2026tiermem} & 8192 & 0 & 8192 & 0.1765 \\
    \name-7B & 4096 & $\le 1243$ & $\le 5339$ & 0.2721 \\
    \name-14B & 4096 & $\le 1516$ & $\le 5612$ & 0.2978 \\
    \bottomrule
  \end{tabular}
  }
\end{table}

\begin{table}[H]
  \centering
  \caption{Reader F1 budget sweep on LongMemEval-RR (turn-level,
  GPT-4o reader, top-$k=10$). Each cell reports a matched reader evaluation
  under the retained source-evidence budget.}
  \label{tab:longmemeval-rr-f1-budget-sweep}
  
  \scriptsize
  \setlength{\tabcolsep}{3.5pt}
  \resizebox{0.94\textwidth}{!}{%
  \begin{tabular}{@{}lccccc@{}}
    \toprule
    \textbf{Method} & \textbf{$B{=}512$} & \textbf{$B{=}1024$} & \textbf{$B{=}2048$} & \textbf{$B{=}4096$} & \textbf{$B{=}8192$} \\
    \midrule
    Recency-$B$ & 0.0012 & 0.0131 & 0.0314 & 0.0645 & 0.0827 \\
    Hybrid salience & 0.0141 & 0.0420 & 0.0556 & 0.1024 & 0.1538 \\
    Source-snippet heuristic & 0.0104 & 0.0575 & 0.0485 & 0.0981 & 0.1763 \\
    TierMem-BudgetRaw~\citep{zhu2026tiermem} & 0.0151 & 0.0353 & 0.0701 & 0.1094 & 0.1765 \\
    Oracle retention & 0.4843 & 0.5013 & 0.5218 & 0.4632 & 0.4499 \\
    \midrule
    \name-7B & 0.2413 & 0.2656 & 0.2715 & 0.2721 & 0.2768 \\
    \name-14B & 0.2314 & 0.2887 & 0.2841 & 0.2978 & 0.3017 \\
    \bottomrule
  \end{tabular}
  }
  
\end{table}

\paragraph{Interpreting the F1 frontier.}
The F1 sweep tracks the same efficiency pattern as the evidence metrics:
\name remains ahead across retained source-evidence budgets and reaches its
strongest reported F1 at the shared $B=8192$ comparison point. Retain-Recall and
Read-Recall in Table~\ref{tab:longmemeval-rr-budget-sweep} separate the two
forces behind this frontier. More budget lets more gold evidence survive
ingestion, but a larger retained cover also gives the fixed top-$k$ retriever
more candidates to rank. The oracle curve makes this visible: oracle retention
nearly saturates Retain-Recall at all budgets, peaks at $B=2048$ with 0.5218 F1,
and then falls to 0.4499 F1 at the shared $B=8192$ anchor as Read-Recall drops
from 0.9219 to 0.8367. Thus 0.4499 is the matched oracle reference for the main
8192-token comparison, not a cross-budget maximum. This non-monotonic oracle
behavior is not a separate anomaly; it is the read-side dilution that motivates
reporting Retain-Recall and Read-Recall separately. The smallest budget remains
informative: \name-7B at $B=512$ reaches 0.2413 F1, already above the strongest
non-\name budgeted baseline at $B=8192$ (0.1765). At this single budget point,
\name-7B has slightly higher F1 than \name-14B even though \name-14B has higher
Retain-Recall and Read-Recall; this is answer-use variance when little context
reaches the reader, not a retention reversal. In source-evidence budget terms,
the same answer pipeline uses 512 instead of 8192 charged source-evidence tokens,
a reduction of 7680 tokens, or about 94\%, per query-time memory state.

\begin{table}[H]
  \centering
  \caption{Offline retention probe on LongMemEval-RR (top-$k=10$).
  The table reports the fair heuristic with the highest Retain-Recall at each
  retained source-evidence budget under the offline probe. Randomized baselines are averaged over five seeds,
  with confidence intervals reported as 95\% CIs when shown; dense baselines use
  deterministic hashing embeddings. Session-level
  rows report oracle retained recall only, so oracle
  retrieved recall is marked n/a. }
  \label{tab:longmemeval-rr-retention-probe}
  
  \scriptsize
  \setlength{\tabcolsep}{3pt}
  \resizebox{\textwidth}{!}{%
  \begin{tabular}{@{}lclcccc@{}}
    \toprule
    \textbf{Granularity} & \textbf{Budget} & \textbf{Best fair heuristic} &
    \textbf{Retain-Recall} & \textbf{Read-Recall} & \textbf{Oracle Retain-Recall} & \textbf{Oracle retrieved (Read-Recall)} \\
    \midrule
    Session & 512 & Source-snippet heuristic & 0.0150 & 0.0150 & 0.0270 & n/a \\
    Session & 1024 & Source-snippet heuristic & 0.0180 & 0.0180 & 0.0810 & n/a \\
    Session & 2048 & TF-IDF salience & 0.0358 & 0.0358 & 0.2790 & n/a \\
    Session & 4096 & Recency-$B$ & 0.0679 & 0.0679 & 0.7069 & n/a \\
    Session & 8192 & Hybrid salience & 0.1297 & 0.1297 & 0.9776 & n/a \\
    \midrule
    Turn & 512 & Source-snippet heuristic & 0.0256 & 0.0231 & 0.9608 & 0.9576 \\
    Turn & 1024 & Source-snippet heuristic & 0.0712 & 0.0658 & 0.9744 & 0.9602 \\
    Turn & 2048 & Source-snippet heuristic & 0.1141 & 0.0812 & 0.9839 & 0.9219 \\
    Turn & 4096 & Source-snippet heuristic & 0.1541 & 0.1258 & 0.9960 & 0.8842 \\
    Turn & 8192 & Source-snippet heuristic & 0.2657 & 0.1754 & 0.9998 & 0.8367 \\
    \bottomrule
  \end{tabular}
  }
  
  \end{table}

\begin{table}[H]
  \centering
  \caption{Hybrid-salience retention probes on LongMemEval-RR.
  Values summarize hybrid-salience retention across single-query and
  multi-query retained-memory settings.}
  \label{tab:hybrid-salience-retention}
  
  \scriptsize
  \setlength{\tabcolsep}{4pt}
  \resizebox{\textwidth}{!}{%
  \begin{tabular}{@{}llccccc@{}}
    \toprule
    \textbf{Setting} & \textbf{Granularity} & \textbf{Budget} &
    \textbf{Top-$k$} & \textbf{Retain-Recall} & \textbf{Read-Recall} &
    \textbf{Coverage balance} \\
    \midrule
    LongMemEval-RR & Turn & 8192 & 10 & 0.2140 & 0.1763 & n/a \\
    LongMemEval-RR & Session & 8192 & 10 & 0.1297 & 0.1297 & n/a \\
    MultiQ-LongMemEval-RR & Group & 10\% & 10 & 0.1063 & 0.1063 & 0.0000 \\
    \bottomrule
  \end{tabular}
  }
  
\end{table}

\begin{table}[H]
  \centering
  \caption{Extended GPT-4o reader evaluation on LongMemEval-RR,
  derived from LongMemEval histories and questions~\citep{wu2025longmemeval}.
  Methods ingest turn-level histories before the query is known, then answer
  after top-$k$ retrieval from either retained memory or full raw logs. Budgeted
  rows use $B=8192$ retained source-evidence tokens unless noted. Full raw-log RAG and
  TierMem-FullRaw keep the full raw history at read time. Oracle retention is an
  oracle reference policy that packs annotated gold-evidence turns into the same
  $B_{\mathrm{ret}}$ before answer-time retrieval; it uses gold
  evidence labels during retention but no answer text at read time. Unknown is
  the fraction of reader outputs that abstain or cannot produce an answer.
  MemAgent rows store
  rewritten memory text; source-unit recall metrics are therefore N/A. \name
  rows report the synchronized F1/Retain/Read
  run used in the main comparison; Sub-EM and Unknown are not reported for those
  rows.}
  \label{tab:longmemeval-rr-reader-check}
  
  \scriptsize
  \setlength{\tabcolsep}{4pt}
  \resizebox{\textwidth}{!}{%
  \begin{tabular}{@{}llcccccc@{}}
    \toprule
    \textbf{Method} & \textbf{Budget} & \textbf{Top-$k$} &
    \textbf{F1} & \textbf{Sub-EM} & \textbf{Unknown} & \textbf{Retain-Recall} & \textbf{Read-Recall} \\
    \midrule
    Full raw-log RAG~\citep{lewis2020retrieval} & full raw & 3 & 0.2567 & 0.2260 & 0.6180 & 1.0000 & 0.3949 \\
    Full raw-log RAG~\citep{lewis2020retrieval} & full raw & 5 & 0.3004 & 0.2640 & 0.5480 & 1.0000 & 0.4814 \\
    Full raw-log RAG~\citep{lewis2020retrieval} & full raw & 10 & 0.3645 & 0.3140 & 0.4340 & 1.0000 & 0.6082 \\
    TierMem-FullRaw~\citep{zhu2026tiermem} & full raw & 10 & 0.4541 & 0.4600 & 0.3820 & 1.0000 & 0.6152 \\
    Oracle retention & 8192 & 3 & 0.3786 & 0.3420 & 0.4680 & 0.9998 & 0.6802 \\
    Oracle retention & 8192 & 5 & 0.4104 & 0.3600 & 0.4100 & 0.9998 & 0.7527 \\
    Oracle retention & 8192 & 10 & 0.4499 & 0.3980 & 0.3680 & 0.9998 & 0.8367 \\
    Random-$B$ & 8192 & 10 & 0.0350 & 0.0129 & 0.9251 & 0.0612 & 0.0425 \\
    Recency-$B$ & 8192 & 10 & 0.0827 & 0.0756 & 0.8756 & 0.1521 & 0.0943 \\
    Reservoir-$B$ & 8192 & 10 & 0.0498 & 0.0275 & 0.9214 & 0.0753 & 0.0615 \\
    Hybrid salience & 8192 & 10 & 0.1538 & 0.1264 & 0.7653 & 0.2140 & 0.1763 \\
    Summary-only & $B$-equiv. & 10 & 0.1156 & 0.0967 & 0.8251 & N/A & N/A \\
    Source-snippet heuristic & 8192 & 10 & 0.1763 & 0.1285 & 0.7852 & 0.2657 & 0.1754 \\
    Generated-query indexing~\citep{nogueira2019doc2query} & 8192 & 10 & 0.1205 & 0.1065 & 0.7402 & 0.2021 & 0.1423 \\
    TierMem-BudgetRaw~\citep{zhu2026tiermem} (Recency-$B$) & 8192 & 10 & 0.1765 & 0.1476 & 0.7126 & 0.1246 & 0.0997 \\
    MemAgent-7B~(pre-query adapted)~\citep{yu2026memagent} & 8192 & 10 & 0.1311 & N/A & 0.8180 & N/A & N/A \\
    \name-7B & 8192 & 10 & 0.2768 & N/A & N/A & 0.2966 & 0.2915 \\
    \name-14B & 8192 & 10 & 0.3017 & N/A & N/A & 0.3215 & 0.3112 \\
    TierMem-BudgetRaw~\citep{zhu2026tiermem} (Recency-$B$) & 8192 & 3 & 0.0503 & 0.0480 & 0.9300 & 0.1246 & 0.0731 \\
    TierMem-BudgetRaw~\citep{zhu2026tiermem} (Dense MMR) & 8192 & 3 & 0.0248 & 0.0180 & 0.9560 & 0.0303 & 0.0208 \\
    TierMem-BudgetRaw~\citep{zhu2026tiermem} (Hybrid salience) & 8192 & 3 & 0.0224 & 0.0200 & 0.9720 & 0.0171 & 0.0130 \\
    TierMem-BudgetRaw~\citep{zhu2026tiermem} (Source-snippet) & 8192 & 3 & 0.0195 & 0.0160 & 0.9700 & 0.0147 & 0.0134 \\
    MemAgent-7B~(query-visible)~\citep{yu2026memagent} & native & 3 & 0.4633 & 0.4700 & 0.4200 & N/A & N/A \\
    \bottomrule
  \end{tabular}
  }
  
\end{table}

\begin{table}[H]
  \centering
  \caption{TierMem-BudgetRaw breakdown on LongMemEval-RR with
  GPT-4o reader, turn-level retention, $B=8192$, and top-$k=3$ query-time
  retrieval. Values report answer F1 by question type.}
  \label{tab:tiermem-budgetraw-breakdown}
  
  \small
  \setlength{\tabcolsep}{5pt}
  \begin{tabular}{@{}lc@{}}
    \toprule
    \textbf{Question type} & \textbf{F1} \\
    \midrule
    Knowledge update & 0.1995 \\
    Multi-session & 0.0030 \\
    Temporal reasoning & 0.0000 \\
    Single-session user & 0.0786 \\
    \bottomrule
  \end{tabular}
  
\end{table}

%
%

\begin{table}[H]
  \centering
  \caption{Diagnostic retained-budget adaptations of existing memory methods.
  Adapted variants ingest the same stream before the query is known, enforce the
  same $B_{\mathrm{ret}}$, and answer only from retained memory. These rows
  use a Qwen2.5-14B reader and are not mixed with the GPT-4o Table~\ref{tab:longmemeval-rr-main}
  comparison.}
  \label{tab:rr-adaptations-existing-memory}
  \scriptsize
  \setlength{\tabcolsep}{3pt}
  \resizebox{\textwidth}{!}{%
  \begin{tabular}{@{}lllllccc@{}}
    \toprule
    \textbf{Method} & \textbf{Original assumption} & \textbf{RR adaptation} & \textbf{Budget enforcement} & \textbf{Raw-log access?} & \textbf{Reader} & \textbf{Retain-Recall} & \textbf{F1} \\
    \midrule
    Memory-R1~\citep{yan2025memoryr1} & Learned memory operations & RR retained-evidence adaptation & Hard retained budget & No & Qwen2.5-14B & 0.1701 & 0.1524 \\
    Mem-$\alpha$~\citep{wang2025memalpha} & QA-reward memory construction & RR retained-evidence adaptation & Hard retained budget & No & Qwen2.5-14B & 0.2341 & 0.1731 \\
    AtomMem~\citep{huo2026atommem} & Atomic memory editing & RR retained-evidence adaptation & Hard retained budget & No & Qwen2.5-14B & 0.1792 & 0.1568 \\
    \bottomrule
  \end{tabular}
  } 
\end{table}

\begin{table}[H]
  \centering
  \caption{Multi-query LongMemEval-RR retention probe. A single
  retained memory is shared by multiple future queries from the same history.
  Values report mean per-query retained evidence recall,
  retrieved evidence recall, and coverage balance under $B_{\mathrm{ret}}$.
  For a query bundle $g$ with queries $q_{g,1:K}$, coverage balance is
  $\min_j \mathrm{RetainRecall}(q_{g,j})$, averaged over bundles; it measures
  whether one frozen memory also covers the hardest query in the bundle.}
  \label{tab:multiq-longmemeval-rr}
  \scriptsize
  \setlength{\tabcolsep}{3pt}
  \resizebox{\textwidth}{!}{%
  \begin{tabular}{@{}cclcccc@{}}
    \toprule
    \textbf{Top-$k$} & \textbf{Budget} & \textbf{Best fair retention} &
    \textbf{Mean query Retain-Recall} & \textbf{Read-Recall} &
    \textbf{Coverage balance} & \textbf{Oracle mean query Retain-Recall} \\
    \midrule
    3 & 1\% & Recency-$B$ & 0.0147 & 0.0147 & 0.0000 & 0.0950 \\
    3 & 2\% & Random-$B$ & 0.0265 & 0.0265 & 0.0000 & 0.1397 \\
    3 & 5\% & Recency-$B$ & 0.0650 & 0.0645 & 0.0000 & 0.3386 \\
    3 & 10\% & Recency-$B$ & 0.1200 & 0.1100 & 0.0000 & 0.6275 \\
    \midrule
    10 & 1\% & Recency-$B$ & 0.0147 & 0.0147 & 0.0000 & 0.0950 \\
    10 & 2\% & Random-$B$ & 0.0265 & 0.0265 & 0.0000 & 0.1397 \\
    10 & 5\% & Recency-$B$ & 0.0650 & 0.0650 & 0.0000 & 0.3386 \\
    10 & 10\% & Recency-$B$ & 0.1200 & 0.1200 & 0.0000 & 0.6275 \\
    \bottomrule
  \end{tabular}
  }
  
\end{table}

\begin{figure}[H]
  \centering
  
  \includegraphics[width=0.62\linewidth]{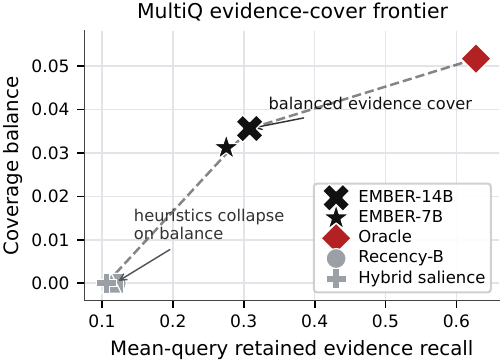}
  
  \caption{MultiQ coverage frontier on LongMemEval-RR. At the same
  10\% of $B_{\mathrm{ret}}$, heuristic retention methods preserve
  fragments of future evidence but fail to cover all future queries.
  \name shifts the retained memory toward a more balanced evidence cover while
  also increasing mean-query evidence recall.}
  \label{fig:multiq-coverage-frontier}
\end{figure}

\begin{table}[H]
  \centering
  \caption{Implementation and training infrastructure used for
  LongMemEval-RR rollouts.}
  \label{tab:rr-implementation-summary}
  
  \scriptsize
  \setlength{\tabcolsep}{5pt}
  \resizebox{\textwidth}{!}{%
  \begin{tabular}{@{}lll@{}}
    \toprule
    \textbf{Component} & \textbf{Implementation choice} & \textbf{Why it matters} \\
    \midrule
    Rollout engine & vLLM rollout server & Supports batched pre-query episodes. \\
    Retriever / index & GPU-resident embedding index & Keeps retrieval latency controlled during budget sweeps. \\
    Episode format & Online ingest followed by post-ingestion queries & Matches the retained-budget protocol. \\
    Budget accounting & Soft over-budget penalty during training; hard retained budget at evaluation & Makes retained evidence cost comparable across methods. \\
    RL interface & Curriculum-selected carrier turn & Applies the PPO/GRPO loss to one configured write or read-selection decision while scoring the full trajectory. \\
    \bottomrule
  \end{tabular}
  }
  
\end{table}

\subsection{\texorpdfstring{Compute, Assets, and Release Notes}{Compute, Assets, and Release Notes}}
\label{app:compute-assets}

Table~\ref{tab:compute-disclosure} summarizes the execution paths
used by the reported experiments. Reader evaluations use GPT-4o through Azure
OpenAI with deterministic decoding. Retention probes and LongMemEval-RR reader
runs share the same retained-budget wrapper and evaluation scripts. The
training and rollout jobs use two GPU configurations: H200 SXM $\times 4$ and
H100 NVL $\times 8$. The
anonymized supplementary artifact includes the construction scripts, budget
settings, evaluation commands, and wrappers needed to reproduce the reported
protocol.

\begin{table}[H]
  \centering
  \caption{Compute disclosure for the reported experiments. Exact
  wall-clock time depends on batching and API throughput; the released scripts
  specify the commands and operating points used for each run.}
  \label{tab:compute-disclosure}
  
  \small
  \setlength{\tabcolsep}{5pt}
  \resizebox{\textwidth}{!}{%
  \begin{tabular}{@{}lll@{}}
    \toprule
    \textbf{Experiment family} & \textbf{Compute path} & \textbf{Reported operating point} \\
    \midrule
    LongMemEval-RR retention sweeps & CPU preprocessing plus embedding retrieval/indexing & $B \in \{512,1024,2048,4096,8192\}$, top-$k=10$ \\
    LongMemEval-RR reader evaluation & Azure GPT-4o reader, temperature $0$ & turn-level, $B=8192$, top-$k=10$ \\
    \name training & vLLM rollout server on H200 SXM $\times 4$ or H100 NVL $\times 8$ & Stage I converges around 500 steps; Stage II/III continuation trains for 100 steps \\
    MultiQ-LongMemEval-RR & Offline retained-evidence probe on CPU/GPU preprocessing path & multi-query retained-budget setting, 1--10\% retained budget \\
    \bottomrule
  \end{tabular}
  }
  
\end{table}

Table~\ref{tab:asset-license-summary} lists the main external assets.
We cite the upstream sources and do not redistribute third-party model weights
or benchmark data beyond derived protocol scripts and evaluation metadata. Users
of the released artifact should obtain each upstream dataset or model under its
own license and terms.

\begin{table}[H]
  \centering
  \caption{External assets used by the paper and how they are used.}
  \label{tab:asset-license-summary}
  
  \small
  \setlength{\tabcolsep}{5pt}
  \resizebox{\textwidth}{!}{%
  \begin{tabular}{@{}lll@{}}
    \toprule
    \textbf{Asset} & \textbf{Use in this paper} & \textbf{License / terms handling} \\
    \midrule
    LongMemEval~\citep{wu2025longmemeval} & Source histories and questions for LongMemEval-RR & Cited upstream; users obtain upstream data under its terms. \\
    RULER / HotpotQA / MuSiQue / 2WikiMultiHopQA~\citep{hsieh2024ruler,trivedi2022musique,ho2020constructing} & Controlled pre-query QA training and stress tests & Cited upstream; benchmark data are not redistributed in the paper. \\
    Qwen2.5 / Qwen3 & Writer and reader backbones & Upstream model sources are credited; third-party weights are not redistributed. \\
    BGE-small~\citep{bge_embedding} & Embedding retriever & Cited upstream; used as an external embedding model. \\
    GPT-4o / Azure OpenAI & Reader evaluation & Used through the hosted API; model weights are not redistributed. \\
    Prior memory systems~\citep{yu2026memagent,yan2025memoryr1,wang2025memalpha,zhu2026tiermem} & Baselines and comparison regimes & Cited upstream; adaptations are evaluated under the retained-budget protocol. \\
    \bottomrule
  \end{tabular}
  }
  
\end{table}

\subsection{\texorpdfstring{Full Rollout Procedure}{Full Rollout Procedure}}
\label{app:full-rollout-procedure}

Algorithm~\ref{alg:appendix-rollout} expands the concise method
description into the full execution procedure used by the rollout engine. The
main text focuses on the retention mechanism; this appendix records the
implementation sequence: online ingest, budgeted cover update, query-time
retrieval from retained memory, evidence selection, answer generation, and
trajectory logging for outcome training.

Here $B_{\text{work}}$ denotes the active-context capacity used to trigger a
write step when the streaming buffer grows too large. It is an implementation
threshold for $w$, not $B_{\mathrm{ret}}$.
All budgeted comparisons vary or fix $B_{\mathrm{ret}}$.

\begin{algorithm}[H]
\caption{\name Rollout with Budgeted Pre-Query Evidence Retention}
\label{alg:appendix-rollout}
\begin{algorithmic}[1]
\Require Stream $c_{1:K}$, query $q$, active-context budget $B_{\text{work}}$, retained evidence budget $B_{\mathrm{ret}}$, initial cover $S_0$, writer policy $\pi^{\mathrm{write}}$, selector $\pi^{\mathrm{sel}}$, reader $\pi^{\mathrm{ans}}$
\Ensure Answer $\hat{y}$, trajectory $\tau$, retained evidence cover $S$
\State $w \gets \emptyset$,\; $S \gets S_0$,\; $\tau \gets ()$
\Statex \hrulefill\; \textit{Phase 1: Pre-query evidence retention} \;\hrulefill
\For{$k = 1$ to $K$}
  \State $w \gets \mathrm{Append}(w, c_k)$
  \While{$|w| > B_{\text{work}}$}
    \State $C^{\text{maint}} \gets \mathrm{LocalMemoryState}(S, w)$
    \State $(M, w') \sim \pi^{\mathrm{write}}(\cdot \mid c_k, w, C^{\text{maint}})$
    \State $S \gets \mathrm{BudgetUpdate}(S, M; B_{\mathrm{ret}})$
    \State $w \gets w'$
    \State Append the memory-control segment and log-probs to $\tau$
  \EndWhile
\EndFor
\Statex \hrulefill\; \textit{Phase 2: Answering from retained evidence} \;\hrulefill
\State $u_{1:n} \sim \pi^{\mathrm{ans}}(\cdot \mid q)$
\State Append $q$ to $u_{1:n}$ if it is not already included
\State $C \gets \mathrm{Retrieve}(S, u_{1:n})$ \Comment{retrieve only from retained evidence}
\State $r \sim \pi^{\mathrm{sel}}(\cdot \mid q, C)$ \Comment{select retained evidence}
\State Append the evidence-selection segment and log-probs to $\tau$
\State $\hat{y} \sim \pi^{\mathrm{ans}}(\cdot \mid q, r)$ \Comment{answer from selected retained evidence}
\State Append solver query and answer segments to $\tau$
\State \Return $\hat{y},\; \tau,\; S$
\end{algorithmic}
\end{algorithm}

\subsection{\texorpdfstring{LongMemEval-S Ablation Settings}{LongMemEval-S Ablation Settings}}
\label{app:longmemeval-ablation-defs}

The LongMemEval-S representation ablations compare memory
representations and interfaces for online ingestion. These settings are not a
fully nested ablation ladder: they test different paths from streamed sessions
to query-time evidence. \textbf{Online notes} asks the LLM to
write generic notes before the query is known, stores the generated note text,
retrieves those notes, and answers from them. This setting represents online
writeable memory with weak evidence grounding: the model may organize the
stream, but it can also discard source details needed later. \textbf{Source
only} stores preserved source snippets directly and retrieves them at query
time. It keeps answer evidence available but lacks retrieval metadata
such as titles, entities, or keys. \textbf{Schema + source} augments preserved
source snippets with retrieval metadata, testing whether source evidence
becomes more useful future memory when paired with an explicit schema. Thus,
  source only is contained in schema + source, but online notes is a separate
  generated-memory path outside the nested source-retention ladder.

\subsection{\texorpdfstring{Pre-Query RULER-HotpotQA Length Sweep}{Pre-Query RULER-HotpotQA Length Sweep}}
\label{app:prequery-length-sweep}

\begin{table}[ht]
  \centering
  \caption{Pre-query RULER-HotpotQA length sweep across input
  lengths up to 3.5M tokens. All methods commit memory or indexes before the
  question is known and see the question only at retrieval or answer time.
  MemAgent rows use the pre-query retraining protocol described in
  Appendix~\ref{app:memagent-prequery}. Because this task family is used in
  Stage I, the table is an in-distribution stress test; LongMemEval-RR is the
  held-out cross-method evaluation. Values report answer F1 on a 0--1 scale.}
  \label{tab:ruler-hqa-prequery-sweep}
  
  \scriptsize
  \setlength{\tabcolsep}{3pt}
  \resizebox{\textwidth}{!}{%
  \begin{tabular}{@{}llcccccccccc@{}}
    \toprule
    \textbf{Model} & \textbf{Method} & \textbf{7K} & \textbf{14K} & \textbf{28K} & \textbf{56K} & \textbf{112K} & \textbf{224K} & \textbf{448K} & \textbf{896K} & \textbf{1.75M} & \textbf{3.5M} \\
    \midrule
    Qwen2.5-7B-Instruct & Context only & 0.6366 & 0.5739 & 0.5237 & 0.0025 & 0.0023 & 0.0000 & 0.0014 & 0.0071 & 0.0022 & 0.0000 \\
    Qwen2.5-7B-Instruct & Vanilla RAG ($k=15$) & 0.7553 & 0.7514 & 0.7484 & 0.7416 & 0.7334 & 0.7279 & 0.7089 & 0.6633 & 0.6525 & 0.6285 \\
    Qwen2.5-7B-Instruct & Vanilla RAG ($k=8$) & 0.7331 & 0.7282 & 0.7308 & 0.7203 & 0.7142 & 0.7124 & 0.7032 & 0.6542 & 0.6401 & 0.6197 \\
    Qwen2.5-14B-Instruct & Context only & 0.7433 & 0.7071 & 0.6447 & 0.0026 & 0.0047 & 0.0000 & 0.0030 & 0.0069 & 0.0025 & 0.0000 \\
    Qwen2.5-14B-Instruct & Vanilla RAG ($k=15$) & 0.7570 & 0.7382 & 0.7369 & 0.7532 & 0.7361 & 0.7395 & 0.6836 & 0.6793 & 0.6395 & 0.5647 \\
    Qwen2.5-14B-Instruct & Vanilla RAG ($k=8$) & 0.7412 & 0.7281 & 0.7458 & 0.7401 & 0.7156 & 0.7215 & 0.6645 & 0.6638 & 0.6196 & 0.5432 \\
    Qwen3-8B & Context only & 0.7671 & 0.7570 & 0.7108 & 0.0026 & 0.0029 & 0.0017 & 0.0039 & 0.0076 & 0.0024 & 0.0000 \\
    Qwen3-8B & Vanilla RAG ($k=15$) & 0.7817 & 0.7685 & 0.7713 & 0.7653 & 0.7584 & 0.7715 & 0.7527 & 0.7429 & 0.6990 & 0.5563 \\
    Qwen3-8B & Vanilla RAG ($k=8$) & 0.7707 & 0.7542 & 0.7772 & 0.7587 & 0.7412 & 0.7653 & 0.7392 & 0.7245 & 0.6775 & 0.5502 \\
    Qwen3-14B & Context only & 0.6880 & 0.6846 & 0.5478 & 0.0075 & 0.0027 & 0.0000 & 0.0087 & 0.0036 & 0.0028 & 0.0000 \\
    Qwen3-14B & Vanilla RAG ($k=15$) & 0.7821 & 0.7627 & 0.7755 & 0.7723 & 0.7603 & 0.7644 & 0.7010 & 0.7312 & 0.6330 & 0.5354 \\
    Qwen3-14B & Vanilla RAG ($k=8$) & 0.8017 & 0.7712 & 0.7542 & 0.7765 & 0.7514 & 0.7556 & 0.7125 & 0.7245 & 0.6225 & 0.4914 \\
    MemAgent-7B & retrained for pre-query writing & 0.2929 & 0.2796 & 0.2606 & 0.2249 & 0.2597 & 0.2665 & 0.2520 & 0.2363 & 0.1500 & 0.2125 \\
    MemAgent-14B & retrained for pre-query writing & 0.2981 & 0.2814 & 0.2069 & 0.2396 & 0.2208 & 0.2494 & 0.2476 & 0.2181 & 0.1631 & 0.1169 \\
    \textbf{\name-7B} & \textbf{Learned memory policy} & 0.8342 & 0.8343 & 0.8195 & \textbf{0.8249} & 0.8116 & 0.8013 & 0.7856 & 0.7907 & 0.7927 & \textbf{0.7856} \\
    \textbf{\name-14B} & \textbf{Learned memory policy} & \textbf{0.8542} & \textbf{0.8613} & \textbf{0.8412} & 0.8214 & \textbf{0.8212} & \textbf{0.8203} & \textbf{0.8023} & \textbf{0.7945} & \textbf{0.8006} & 0.7851 \\
    \bottomrule
  \end{tabular}
  }
  
\end{table}

%
%

\subsection{\texorpdfstring{Question-aware RULER-HotpotQA Accuracy}{Question-aware RULER-HotpotQA Accuracy}}
\label{app:length-sweep}

QwenLong-L1 is reported here under its native question-aware
long-context protocol because it does not expose a separate memory-construction
step to which pre-query writing can be applied.

\begin{table}[ht]
  \centering
  \caption{Question-aware RULER-HotpotQA results across input
  lengths up to 3.5M tokens. This prior-work protocol is separate from both
  pre-query evaluation and LongMemEval-RR retained-budget evaluation: the
  model has access to the question while processing the input. Values report
  prior-work accuracy (\%), not answer F1. MemAgent and QwenLong-L1 rows are
  reported by MemAgent~\citep{yu2026memagent}. N/A denotes lengths not reported
  or not supported under the corresponding prior-work setting.}
  \label{tab:ruler-length-sweep}
  
  \scriptsize
  \setlength{\tabcolsep}{3pt}
  \resizebox{\textwidth}{!}{%
  \begin{tabular}{@{}llcccccccccc@{}}
    \toprule
    \textbf{Model} & \textbf{Method} & \textbf{7K} & \textbf{14K} & \textbf{28K} & \textbf{56K} & \textbf{112K} & \textbf{224K} & \textbf{448K} & \textbf{896K} & \textbf{1.75M} & \textbf{3.5M} \\
    \midrule
    DeepSeek-R1-Distill-Qwen-7B & Context only & 32 & 12.50 & 3.12 & 0.71 & 0.62 & 1.27 & 0.21 & N/A & N/A & N/A \\
    DeepSeek-R1-Distill-Qwen-7B & Vanilla RAG ($k=6$) & 61.13 & 54.87 & 53.17 & 49.23 & 50.32 & 32.43 & 33.85 & 30.08 & 29.21 & 21.65 \\
    DeepSeek-R1-Distill-Qwen-14B & Context only & 64.06 & 64.84 & 57.03 & 40.62 & 14.84 & 8.59 & 3.12 & 6.25 & N/A & N/A \\
    DeepSeek-R1-Distill-Qwen-14B & Vanilla RAG ($k=6$) & 69.53 & 70.12 & 71.41 & 46.27 & 45.56 & 42.34 & 37.24 & 32.34 & 33.17 & 30.19 \\
    Qwen2.5-7B-Instruct & Context only & 55.45 & 56.75 & 53.78 & 50.51 & 47.45 & 26.34 & 7.25 & N/A & N/A & N/A \\
    Qwen2.5-7B-Instruct & Vanilla RAG ($k=6$) & 72.13 & 70.34 & 70.13 & 69.23 & 67.45 & 62.76 & 61.90 & 60.52 & 60.21 & 59.35 \\
    Qwen2.5-14B-Instruct & Context only & 70.76 & 68.19 & 56.78 & 53.2 & 43.23 & 33.55 & 11.25 & 3.56 & N/A & N/A \\
    Qwen2.5-14B-Instruct & Vanilla RAG ($k=6$) & 75.64 & 74.34 & 75.14 & 74.12 & 69.53 & 69.67 & 71.74 & 68.54 & 66.32 & 67.79 \\
    Qwen3-8B & Context only & 57.45 & 60.43 & 55.78 & 54.51 & 50.45 & 31.34 & 11.25 & 1.54 & N/A & N/A \\
    Qwen3-8B & Vanilla RAG ($k=6$) & 70.69 & 72.54 & 71.31 & 70.85 & 68.45 & 67.62 & 67.18 & 64.83 & 64.49 & 63.78 \\
    Qwen3-14B & Context only & 72.23 & 69.10 & 59.28 & 55.17 & 50.82 & 45.09 & 22.13 & 15.42 & N/A & N/A \\
    Qwen3-14B & Vanilla RAG ($k=6$) & 75.64 & 74.34 & 75.14 & 74.12 & 69.53 & 69.67 & 71.74 & 68.54 & 66.32 & 67.79 \\
    QwenLong-L1-32B & Long-context RL model & 72.66 & 75.00 & 72.66 & 60.94 & 31.25 & 17.19 & 13.28 & 11.72 & N/A & N/A \\
    MemAgent-7B & RL in-context memory & 81.63 & 80.42 & 79.23 & 78.16 & 79.73 & 73.84 & 74.56 & 75.18 & 74.97 & 70.58 \\
    MemAgent-14B & RL in-context memory & 83.69 & 83.23 & 84.54 & 80.86 & 77.68 & 80.73 & 74.90 & 78.03 & 76.43 & \textbf{77.09} \\
    \textbf{\name-7B} & \textbf{Learned memory policy} & 84.32 & \textbf{86.53} & \textbf{85.43} & 85.67 & 85.17 & \textbf{84.78} & \textbf{83.05} & 80.42 & \textbf{78.56} & 76.67 \\
    \textbf{\name-14B} & \textbf{Learned memory policy} & \textbf{85.64} & 85.42 & 85.08 & \textbf{86.32} & \textbf{86.54} & 83.65 & 81.87 & \textbf{81.52} & 76.61 & 76.14 \\
    \bottomrule
  \end{tabular}
  }
  
\end{table}

\subsubsection{\texorpdfstring{W0 Diagnostic Probe}{W0 Diagnostic Probe}}
\label{app:w0-diagnostic}

To localize the bottleneck of the trained policy, we use a fixed-action
diagnostic rollout with the trained checkpoint. During rollout, each memory write
is instrumented with provenance
tags and measured whether each stored item was later recovered by retrieval,
whether it was selected into the recall context, and whether it contained the
gold evidence span. Table~\ref{tab:w0-diagnostic} reports the resulting
stage-wise metrics.

\begin{table}[ht]
  \centering
  \scriptsize
  \setlength{\tabcolsep}{3pt}
  \caption{W0 training diagnostic rollout on held-out episodes,
  reported separately from the LongMemEval-RR retained-budget benchmark.
  Retrieval recall is saturated under the default top-$k$ setting, while losses
  appear in write quality and read-side selection.}
  \label{tab:w0-diagnostic}
  \begin{tabular}{@{}p{0.23\linewidth}cp{0.38\linewidth}@{}}
    \toprule
    \textbf{Stage} & \textbf{Value} & \textbf{Interpretation} \\
    \midrule
    Written evidence coverage & 47.8\% & About half of written items contain gold evidence. \\
    Retrieval hit rate & 100.0\% & The retriever recovers written memory at default top-$k$. \\
    Read-selection hit rate & 41.7\% & Read selection keeps a minority of written items. \\
    Answer-context gold rate & 77.3\% & Some answers still lack selected gold evidence. \\
    \bottomrule
  \end{tabular}
\end{table}

The diagnostic shows that retrieval recall is not the current
bottleneck: every stored item was recovered by the retriever under the default
top-$k$ setting. The remaining loss comes from write quality and read-side
selection. Only 47.8\% of written items contain the gold evidence span, only
41.7\% of written items are selected into the recall context, and 22.7\% of
episodes still reach the answer stage without selected gold evidence. This
points to write quality and read-side selection; retrieval capacity alone is not
the bottleneck.

\subsection{\texorpdfstring{Training and Reward Details}{Training and Reward Details}}
\label{app:training-config}

Algorithm~\ref{alg:grpo} in Section~\ref{sec:training} summarizes the training
rollout and group-relative update. This appendix gives the instrumentation,
episode construction, coefficient selection, and token-level objective used by
that algorithm.

The auxiliary reward terms in Section~\ref{sec:training} are computed
from rollout instrumentation. Evidence coverage $E_i$ measures whether the
stored or selected evidence contains gold support. The rank score
$L_i$ rewards placing gold evidence high in the retrieved set.
Selection purity $P_i$ measures how much of the selected evidence is useful
and not distractor text. Write utility $W_i$ rewards valid memory writes that preserve
future-use evidence. All auxiliary terms are normalized to $[0,1]$ before being
gated by final answer quality $Q_i$. The budget penalty is
$\lambda_{\mathrm{budget}}\max(0, |S_i|_{\mathrm{tok}}-\tilde{B}_i)/\tilde{B}_i$, which charges
the policy for source evidence retained beyond the sampled budget.
For the reported runs we use
$\lambda_{\mathrm{budget}}=0.2$, rollout group size $G=16$, sampling
temperature $T=1.0$, and source-excerpt caps of 256 tokens with a 128-token
source prefix in the retrieval index.

\paragraph{Training data construction.}
The curriculum trains the retention interface, not a standalone
reader. It has two optimizer-length blocks. Stage I uses RULER-HotpotQA to
create controlled long-context streams with support evidence and distractors and
converges in about 500 optimization steps. Stage II is a 100-step multi-session
continuation: it converts MuSiQue and 2WikiMultiHopQA examples into episodes by
assigning support facts to timestamped sessions, inserting distractor sessions,
and preserving support unit identifiers for reward computation. Stage III hard
cases are folded into this 100-step continuation; they are not a separate
optimizer block. These transformations add update, temporal, and abstention
episodes. In update episodes, an early session states an old value and a later
session corrects it; the writer must preserve the evidence that determines the
current answer. In temporal episodes, the query depends on when an event,
deadline, or preference holds. In abstention episodes, the stream contains
topically related content but no source evidence sufficient to answer the query.
These cases train the writer to prefer gold source evidence over merely topical
context. LongMemEval
histories, questions, and evidence labels are not used for training; all
LongMemEval-derived protocols are held out for evaluation.

\paragraph{Reward coefficient selection.}
We select
$(\alpha_Q,\alpha_E,\alpha_L,\alpha_P,\alpha_W) =
(0.45,\,0.25,\,0.15,\,0.10,\,0.05)$ on a held-out validation split constructed
from the external training sources: RULER-HotpotQA, MuSiQue, and
2WikiMultiHopQA-derived episodes. These validation episodes are disjoint from
training episodes and from the LongMemEval-RR and MultiQ-LongMemEval-RR
evaluation sets. All reported results use this frozen coefficient vector.

\begin{table}[htb]
  \centering
  \caption{Reward-coefficient sensitivity on LongMemEval-RR for \name-7B at
  $B=8192$. The default coefficient vector is selected on external validation
  episodes; the alternatives vary the reward emphasis without test-set tuning.}
  \label{tab:reward-weight-sensitivity}
  \small
  \setlength{\tabcolsep}{5pt}
  \begin{tabular}{@{}lccc@{}}
    \toprule
    \textbf{Reward coefficients $(\alpha_Q,\alpha_E,\alpha_L,\alpha_P,\alpha_W)$} &
    \textbf{Retain-Recall $\uparrow$} &
    \textbf{Read-Recall $\uparrow$} &
    \textbf{F1 $\uparrow$} \\
    \midrule
    $(0.45,\,0.25,\,0.15,\,0.10,\,0.05)$ & \textbf{0.2966} & \textbf{0.2915} & \textbf{0.2768} \\
    $(0.50,\,0.20,\,0.15,\,0.10,\,0.05)$ & 0.2917 & 0.2856 & 0.2733 \\
    $(0.40,\,0.30,\,0.15,\,0.10,\,0.05)$ & 0.2989 & 0.2854 & 0.2722 \\
    All equal: $(0.20,\,0.20,\,0.20,\,0.20,\,0.20)$ & 0.2876 & 0.2711 & 0.2552 \\
    \midrule
    Top non-\name budgeted F1 baseline & 0.1246 & 0.0997 & 0.1765 \\
    \bottomrule
  \end{tabular}
\end{table}

\begin{table}[htb]
  \centering
  \caption{Training-seed stability for \name-7B on LongMemEval-RR at
  $B=8192$. Each seed uses the same reward coefficients, training curriculum,
  budget sampling, and checkpoint selection on held-out external pre-query
  validation episodes. LongMemEval-RR is used only for final evaluation. We use
  \name-7B as the lower-cost replicated setting for estimating training-seed
  variance; \name-14B is reported as a single-seed main model.}
  \label{tab:training-seed-stability}
  \small
  \setlength{\tabcolsep}{6pt}
  \begin{tabular}{@{}lccc@{}}
    \toprule
    \textbf{Seed} &
    \textbf{Retain-Recall $\uparrow$} &
    \textbf{Read-Recall $\uparrow$} &
    \textbf{F1 $\uparrow$} \\
    \midrule
    0 & 0.2966 & 0.2915 & 0.2768 \\
    1 & 0.2912 & 0.2856 & 0.2705 \\
    2 & 0.2954 & 0.2926 & 0.2801 \\
    \midrule
    Mean $\pm$ std & $0.2944{\pm}0.0028$ & $0.2899{\pm}0.0038$ & $0.2758{\pm}0.0049$ \\
    \midrule
    TierMem-BudgetRaw (best F1 baseline) & 0.1246 & 0.0997 & 0.1765 \\
    \bottomrule
  \end{tabular}
\end{table}

\begin{table}[htb]
  \centering
  \caption{Reward-term ablation on LongMemEval-RR for \name-7B at $B=8192$.
  All rows use the same pre-query retention protocol and reader. The answer-only
  RL row keeps reinforcement learning but removes the auxiliary evidence-chain
  terms; it is distinct from the no-RL ablation in Table~\ref{tab:longmemeval-rr-ablation-main}.
  The remaining rows remove retained-evidence coverage $E$, lookup/readability
  score $L$, and selection purity $P$.}
  \label{tab:reward-weight-ablation}
  \small
  \setlength{\tabcolsep}{5pt}
  \begin{tabular}{@{}lccc@{}}
    \toprule
    \textbf{Reward} &
    \textbf{Retain-Recall $\uparrow$} &
    \textbf{Read-Recall $\uparrow$} &
    \textbf{F1 $\uparrow$} \\
    \midrule
    \textbf{Default} & \textbf{0.2966} & \textbf{0.2915} & \textbf{0.2768} \\
    Answer-only RL reward & 0.2666 & 0.2415 & 0.2361 \\
    w/o E & 0.2556 & 0.2416 & 0.2354 \\
    w/o L & 0.2712 & 0.2448 & 0.2322 \\
    w/o P & 0.2759 & 0.2571 & 0.2425 \\
    \midrule
    \bottomrule
  \end{tabular}
\end{table}

The sweep shows that \name is stable under small coefficient changes. Nearby
answer-heavy and evidence-heavy variants stay within 0.005 F1 of the reported
configuration, while equal weighting is lower. We therefore use the rounded
validation-selected vector for all reported runs.

Table~\ref{tab:reward-weight-ablation} ablates the same objective by removing
individual reward signals. The default objective gives the strongest full-chain
performance. Answer-only RL loses both retention and read-time access, while
removing $E$, $L$, or $P$ damages different points of the
Survive--Read--Answer chain. This supports the training design used in the main
experiments: final answer quality gates the reward, but evidence-retention and
readability signals are needed to assign credit to pre-query memory decisions.

For each training batch, we sample a budget
$B_{\mathrm{ret}}\in\{512,1024,2048,4096,8192\}$ for each episode, so training
covers more than the 8192-token operating point. This exposes the writer to the
same retained-memory frontier used in Figure~\ref{fig:budget-frontier}: the
policy must learn which evidence survives when the budget is tight as well as
when the cover is larger.

\paragraph{Validation and checkpoint selection.}
Checkpoint selection uses held-out pre-query validation episodes
constructed from the same external sources and transformations as training.
LongMemEval-RR and MultiQ-LongMemEval-RR are used only after checkpoint
selection, so the main external evaluation does not tune the retention policy on
LongMemEval histories, questions, or support labels.

\paragraph{Episode and batch format.}
Training and evaluation share a single episode schema. Each episode
contains the pre-query stream, the hidden query, the target answer, annotated
support units when available, $B_{\mathrm{ret}}$, and a coarse task
type:

\begin{PromptBlock}
{
  "stream": [
    {"session_id": 1, "timestamp": "...", "text": "..."},
    {"session_id": 2, "timestamp": "...", "text": "..."}
  ],
  "hidden_query": "...",
  "answer": "...",
  "support_units": [...],
  "budget": 512 | 1024 | 2048 | 4096 | 8192,
  "task_type": "single-hop | multi-hop | update | temporal | abstention"
}
\end{PromptBlock}

For the GRPO-style update in Section~\ref{sec:training}, the
old-policy importance ratio for an updated token is
\[
\rho_{i,t}(\theta)=
\frac{\pi_\theta(o_{i,t}\mid \mathrm{ctx}_{i,<t})}
     {\pi_{\theta_{\mathrm{old}}}(o_{i,t}\mid \mathrm{ctx}_{i,<t})}.
\]
For the exposed memory-control tokens $\mathcal{T}_i$, we use the clipped
objective with $\epsilon=0.2$ and $\beta=0.001$:
\[
\mathcal{J}(\theta)=
\mathbb{E}_{i,t\in\mathcal{T}_i}\!\left[
\min\!\Big(\rho_{i,t}(\theta)\hat{A}_i,\,
\mathrm{clip}\!\big(\rho_{i,t}(\theta),1-\epsilon,1+\epsilon\big)\hat{A}_i\Big)
-\beta\,\mathrm{KL}\!\big(\pi_\theta\|\pi_{\mathrm{ref}}\big)
\right].
\]
The reported veRL integration exposes one carrier turn per rollout to the
PPO/GRPO update, while the reward is computed from the full memory trajectory.
The curriculum determines which decision becomes the carrier. In write-retention
batches, the carrier is a writer action block produced during ingestion, such as
a capsule proposal, update mode, or bounded source-excerpt decision for one
chunk window. The rest of the episode is still executed: the query is revealed,
retained evidence is retrieved and selected, and the final answer is scored.
Thus the selected write action receives delayed outcome credit from the full
Survive--Read--Answer chain. Across training batches, different write decisions
are selected as carriers, covering the stream's per-chunk retention decisions.
When the read-selection
curriculum is enabled, the carrier is instead the evidence-selection turn; write
and answer turns are masked from the loss but remain in the trajectory used to
compute the reward. The answer turn is always score-only in these runs. This
applies standard clipped GRPO to a transparent decision carrier while avoiding
token-level credit over every prompt and answer token in a rollout.

\begin{table}[ht]
  \centering
  \caption{Training configuration for the reported \name runs.}
  \label{tab:training-config}
  
  \small
  \setlength{\tabcolsep}{5pt}
  \begin{tabular}{@{}lp{0.62\linewidth}@{}}
    \toprule
    \textbf{Setting} & \textbf{Value} \\
    \midrule
    Stage I data & RULER-HotpotQA pre-query online-memory episodes \\
    Stage II data & MuSiQue and 2WikiMultiHopQA multi-session retention episodes \\
    Stage III data & Update, temporal, and abstention hard-case transformations \\
    Stage I convergence & about 500 optimization steps \\
    Stage II/III continuation & 100 optimization steps \\
    Learning rate & $1\times 10^{-6}$ \\
    Rollout group size & $G=16$ \\
    Rollout sampling temperature & $T=1.0$ \\
    Source excerpt cap & 256 tokens \\
    Index source-prefix cap & 128 tokens \\
    Budget penalty coefficient & $\lambda_{\mathrm{budget}}=0.2$ \\
    PPO clip range & $\epsilon=0.2$ \\
    PPO epochs & 8 \\
    KL coefficient & $\beta=0.001$ \\
    Trainable carrier & Curriculum-selected writer action block or read-selection turn; one carrier receives loss per rollout \\
    Optimizer & Answer-gated evidence-chain reward with clipped GRPO-style update \\
    \bottomrule
  \end{tabular}
  
\end{table}

\clearpage
\section*{NeurIPS Paper Checklist}

\providecommand{\answerYes}{\textbf{[Yes]}}
\providecommand{\answerNo}{\textbf{[No]}}
\providecommand{\answerNA}{\textbf{[N/A]}}

\begin{enumerate}[leftmargin=1.35em]
  \item \textbf{Claims}

  \textbf{Question:} Do the main claims made in the abstract and introduction accurately reflect the paper's contributions and scope?

  \textbf{Answer:} \answerYes

  \textbf{Justification:} The abstract and introduction state the setting, method, and measured gains under Budgeted Pre-Query Retention. The main quantitative claims are reported in Table~\ref{tab:longmemeval-rr-main}, Figure~\ref{fig:budget-frontier}, and Section~\ref{sec:experiments}.

  \item \textbf{Limitations}

  \textbf{Question:} Does the paper discuss the limitations of the work performed by the authors?

  \textbf{Answer:} \answerYes

  \textbf{Justification:} Section~\ref{sec:limitations-impact} discusses the benchmark scope, annotation and budget assumptions, and deployment risks for persistent agent memory.

  \item \textbf{Theory assumptions and proofs}

  \textbf{Question:} For each theoretical result, does the paper provide the full set of assumptions and a complete and correct proof?

  \textbf{Answer:} \answerNA

  \textbf{Justification:} The paper is empirical and does not present formal theoretical results.

  \item \textbf{Experimental result reproducibility}

  \textbf{Question:} Does the paper fully disclose all the information needed to reproduce the main experimental results of the paper to the extent that it affects the main claims and/or conclusions of the paper?

  \textbf{Answer:} \answerYes

  \textbf{Justification:} Section~\ref{sec:method} describes the retention policy, Section~\ref{sec:training} gives the training objective, and Appendix~\ref{app:longmemeval-rr} defines the LongMemEval-RR protocol, metrics, budgets, and evaluation wrapper.

  \item \textbf{Open access to data and code}

  \textbf{Question:} Does the paper provide open access to the data and code, with sufficient instructions to faithfully reproduce the main experimental results, as described in supplemental material?

  \textbf{Answer:} \answerYes

  \textbf{Justification:} Appendix~\ref{app:longmemeval-rr} specifies the derived-protocol construction, retained-budget settings, chunk/span rules, answer-time wrapper, and evaluator. Appendix~\ref{app:compute-assets} states that the anonymized supplementary artifact includes the construction scripts, evaluation commands, and wrappers.

  \item \textbf{Experimental setting/details}

  \textbf{Question:} Does the paper specify all the training and test details necessary to understand the results?

  \textbf{Answer:} \answerYes

  \textbf{Justification:} Section~\ref{sec:experiments} describes the benchmark hierarchy, main baselines, and reader settings. Appendix~\ref{app:training-config} reports the training configuration, and Appendix~\ref{app:longmemeval-rr} reports the protocol and extended tables.

  \item \textbf{Experiment statistical significance}

  \textbf{Question:} Does the paper report error bars suitably and correctly defined or other appropriate information about the statistical significance of the experiments?

  \textbf{Answer:} \answerYes

  \textbf{Justification:} Main F1 tables define uncertainty in their captions: LongMemEval-RR reports bootstrap confidence half-widths, while the RULER-HotpotQA headline table reports 95\% confidence-interval half-widths. The primary LongMemEval-RR gain reports a paired bootstrap confidence interval. Randomized retention probes average five seeds; extended deterministic diagnostics report point estimates.

  \item \textbf{Experiments compute resources}

  \textbf{Question:} For each experiment, does the paper provide sufficient information on the computer resources needed to reproduce the experiments?

  \textbf{Answer:} \answerYes

  \textbf{Justification:} Appendix Table~\ref{tab:rr-implementation-summary} summarizes the rollout engine, GPU-resident retriever, hard budget enforcement, and training interface. Appendix Table~\ref{tab:compute-disclosure} reports the compute paths, operating points, and GPU configurations, including H200 SXM $\times 4$ and H100 NVL $\times 8$.

  \item \textbf{Code of ethics}

  \textbf{Question:} Does the research conducted in the paper conform, in every respect, with the NeurIPS Code of Ethics?

  \textbf{Answer:} \answerYes

  \textbf{Justification:} The work uses public benchmarks and standard language-model evaluation protocols. It does not involve deception, private user-data collection, or human-subject intervention.

  \item \textbf{Broader impacts}

  \textbf{Question:} Does the paper discuss both potential positive societal impacts and negative societal impacts of the work performed?

  \textbf{Answer:} \answerYes

  \textbf{Justification:} Section~\ref{sec:limitations-impact} discusses both the positive impact of reducing unnecessary memory/context and the risks of persistent source-evidence retention, including sensitive-history retention and user control.

  \item \textbf{Safeguards}

  \textbf{Question:} Does the paper describe safeguards that have been put in place for responsible release of data or models that have a high risk for misuse?

  \textbf{Answer:} \answerNA

  \textbf{Justification:} The paper does not release a new general-purpose pretrained language model or a high-risk scraped dataset.

  \item \textbf{Licenses for existing assets}

  \textbf{Question:} Are the creators or original owners of assets used in the paper properly credited and are the license and terms of use explicitly mentioned and properly respected?

  \textbf{Answer:} \answerYes

  \textbf{Justification:} Appendix Table~\ref{tab:asset-license-summary} lists the main external assets, how they are used, and how upstream licenses or service terms are handled. The paper cites the creators of the datasets, model families, retrieval components, and baseline methods used in training and evaluation.

  \item \textbf{New assets}

  \textbf{Question:} Are new assets introduced in the paper well documented and is the documentation provided alongside the assets?

  \textbf{Answer:} \answerYes

  \textbf{Justification:} Appendix~\ref{app:longmemeval-rr} documents LongMemEval-RR, including its source data, construction rules, retained-budget settings, online ingest protocol, metrics, and evaluation wrapper. Appendix~\ref{app:compute-assets} describes the anonymized supplementary artifact and upstream asset handling.

  \item \textbf{Crowdsourcing and research with human subjects}

  \textbf{Question:} For crowdsourcing experiments and research with human subjects, does the paper include the full text of instructions given to participants and screenshots, if applicable, as well as details about compensation?

  \textbf{Answer:} \answerNA

  \textbf{Justification:} The work does not conduct crowdsourcing or human-subject experiments.

  \item \textbf{Institutional review board approvals or equivalent for research with human subjects}

  \textbf{Question:} Does the paper describe potential risks incurred by study participants, whether such risks were disclosed to the subjects, and whether Institutional Review Board approvals were obtained?

  \textbf{Answer:} \answerNA

  \textbf{Justification:} The work does not conduct human-subject research.

  \item \textbf{Declaration of LLM usage}

  \textbf{Question:} Does the paper describe the usage of LLMs if it is an important, original, or non-standard component of the core methods in this research?

  \textbf{Answer:} \answerYes

  \textbf{Justification:} The paper describes the LLM-based memory writer, reader, retrieval prompts, Qwen backbones, GPT-4o reader evaluation, and prompt templates in the method, experiments, and appendices.
\end{enumerate}

\end{document}